\documentclass[10pt,journal,compsoc]{IEEEtran}

\usepackage[utf8]{inputenc}
\usepackage[english]{babel}
\usepackage[export]{adjustbox}

\usepackage{amsthm}
\usepackage{graphicx}
\usepackage{caption,subcaption}
\usepackage{subcaption}
\usepackage{subfloat}
\usepackage{enumitem}
\usepackage{amsmath}
\usepackage{algorithm}
\usepackage{xcolor}
\usepackage[noend]{algpseudocode}

\makeatletter
\def\BState{\State\hskip-\ALG@thistlm}
\makeatother

\usepackage{algorithm}
\usepackage[noend]{algpseudocode}

\algnewcommand\algorithmicinput{\textbf{Input:}}
\algnewcommand\INPUT{\item[\algorithmicinput]}

\algnewcommand\algorithmicoutput{\textbf{Output:}}
\algnewcommand\OUTPUT{\item[\algorithmicoutput]}

\begin{document}





%
\title{Time Series Clustering for Human Behavior Pattern Mining}

%
%
%
%
%

\author{Rohan~Kabra\textsuperscript{*},
        Divya~Saxena\textsuperscript{*},~\IEEEmembership{Member,~IEEE,}
        Dhaval~Patel,~\IEEEmembership{Senior Member,~IEEE,}
        and~Jiannong~Cao,~\IEEEmembership{Fellow,~IEEE}
\IEEEcompsocitemizethanks{
\IEEEcompsocthanksitem *Authors contributed equally.
\IEEEcompsocthanksitem R. Kabra was with the Department of CSE, IIT Roorkee, Roorkee, India.\protect\\
E-mail: rrk.kab05@gmail.com
\IEEEcompsocthanksitem D. Saxena is  with Department of Computing), The Hong Kong Polytechnic University, Hong Kong. E-mail: divsaxen@comp.polyu.edu.hk
\IEEEcompsocthanksitem D. Patel is with IBM T.J. Watson Research Center, NY, USA. E-mail: pateldha@us.ibm.com
\IEEEcompsocthanksitem J. Cao is with Department of Computing, The Hong Kong Polytechnic University, Hong Kong. E-mail: csjcao@comp.polyu.edu.hk}
\thanks{Manuscript received Month, DD Year; revised Month, DD Year.}}

\markboth{Journal of \LaTeX\ Class Files,~Vol.~14, No.~8, August~2015}%
{Shell \MakeLowercase{\textit{et al.}}: Bare Demo of IEEEtran.cls for Computer Society Journals}
%

\IEEEtitleabstractindextext{%
\begin{abstract}
Human behavior modeling deals with learning and understanding of behavior patterns inherent in humans' daily routine. Existing pattern mining techniques either assume human dynamics is strictly periodic, or require the number of modes as input, or do not consider uncertainty in the sensor data. To handle these issues, in this paper, we propose a novel clustering approach for modeling human behavior (named, MTpattern) from time-series data. For mining frequent human behavior patterns effectively, we utilize three-stage pipeline: (1) represent time series data into sequence of regularly sampled equal sized unit time intervals for better  analysis, (2) a new distance measure scheme is proposed to cluster similar sequences which can handle temporal variation and uncertainty in the data, and (3) exploit an exemplar-based clustering mechanism and fine-tune its parameters to output minimum number of clusters with given permissible distance constraints and without knowing the number of modes present in the  data. Then, the average of all sequences in a cluster is considered as a human behavior pattern. Empirical studies on two real-world datasets and a simulated dataset demonstrate the effectiveness of \textit{MTpattern} w.r.to internal and external measures of clustering.

\end{abstract}

}

\maketitle
\IEEEdisplaynontitleabstractindextext

%
\IEEEpeerreviewmaketitle
\IEEEraisesectionheading{\section{Introduction}\label{sec:introduction}}
\IEEEPARstart{M}odeling human behavior using data originating from social network, Internet of Things (IoT), smart home is an active area of research. A behavior pattern refers to a recurrent way of acting or conduct by an individual or a group, such as mobile devices, animals, vehicles, etc., in a physical/virtual environment, while learning and understanding human behavior patterns from raw data is known as \textit{human behavior modeling}. Researchers have discovered and identified various types of behaviors on the basis of the source and domain of raw data, such as social behavior using online social networks \cite{LongOnlineSocial,Xiang:2010:MRS:1772690.1772790}, biological health behavior using smart body sensors \cite{StefanEnvHealth}, online user behavior analysis using clickstream data \cite{unsupervised2016}, customer energy  consumption behavior \cite{learning2018}, customer spending behavior using transaction data \cite{purtreeclust2017}, human activity behavior using multivariate temporal data \cite{multivariate2016} and mobility behavior using smart card data, GPS and WiFi traces \cite{clustering2016, RenTSOS15, scalable2018}, etc. Identifying frequent human behaviors is very challenging in any behavior analysis. The complexity of modeling human behaviors comes from two aspects: vast types of humans and irregular behaviors of each human type. Mining patterns in human behavior have wide practical applications in several domains, such as recommendation systems, healthcare, transportation, etc. For instance, identifying underlying patterns in human moving behavior in a location, such as mall, restaurant, during the COVID-19 pandemic (i.e., contact tracing) can support the authorities to understand who and how many people were in close contact with each other.

\begin{figure}[htbp]
\centerline{\includegraphics[scale= 0.45]{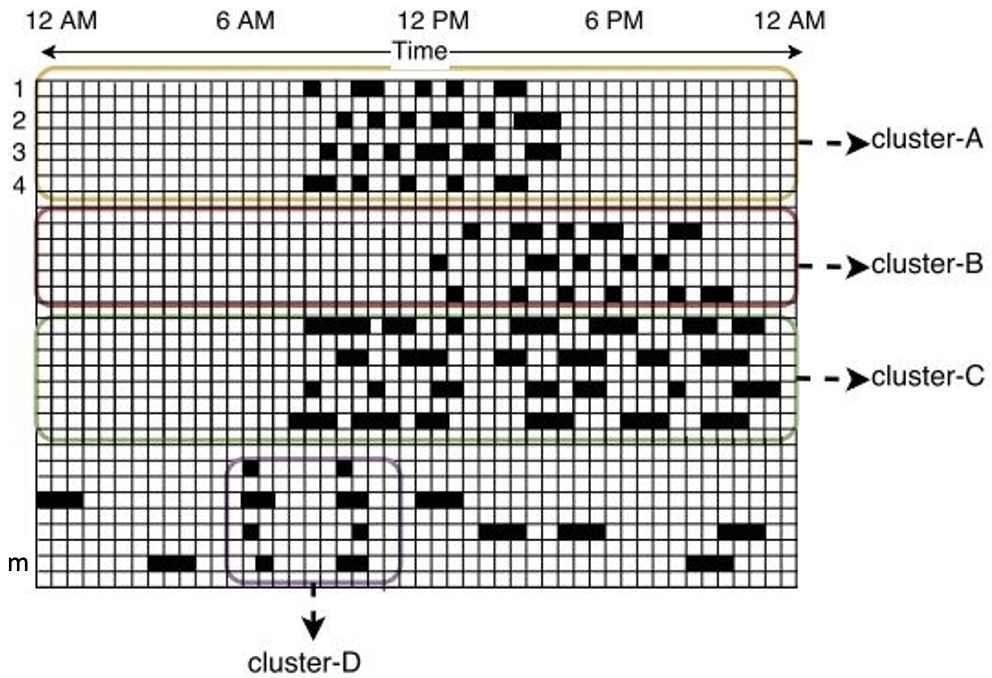}}
\caption{Observed visiting sequences of $m$ individuals.}
\label{fig:sensor_visiting_sequence}
\end{figure}

The main aim of the existing human visiting (or, mobility) behavior pattern mining techniques is to model humans/individuals behavior at single or group level in temporal, spatial or spatio-temporal dimension. Some of the popular behavior pattern mining techniques, include frequency spectrum analysis \cite{ldh10} to analyze the periodicity of recurring behavior patterns, partition and model based time-series clustering \cite{ldh10,WarrenLiao20051857,Bicego2003,Zhong,ICML2011Li159,KlForHMM} and PCA \cite{NathanEigStructure} based eigenbehavior technique to extract the structure of behavior patterns. 

Even though existing  pattern mining methods have shown potential for mining visiting behavior patterns, they have the following limitations: (1) assume temporal  dynamics  is  strictly  periodic. However, in practice, human behavior patterns are temporally variable inherently; (2) The sensor is not accurate and may produce false negatives (non-deterministic values). This leads to uncertainty in data which has not been addressed yet; (3) Multi-modality is inherent in behavior of individuals which means that multiple patterns may occur during a given time window, but number of those patterns are unknown. The pattern may span an entire day or for a short span of the day. This poses a challenge to identify the time of occurrence and duration of all patterns. Most of the techniques need the number of modes of behavior as input and only identify patterns that span the entire time period. 

In many real-world problems, objects are described by large number of binary features. For instance, documents are characterized by presence or absence of certain keywords \cite{santra2016bayesian}; cancer patients are characterized by presence or absence of certain mutations \cite{BioNumerics}; traffic road accident and crash patterns are identified using the presence or absence of accidents and crashes data \cite{rahimi2019clustering}\cite{chung2013identifying}; trading patterns are identified by analysing the presence or the absence of a trading activity \cite{fokianos2017binary};

To better understand the attributes and characteristics of data used to find the behavior of individuals to a location, we take an example. Suppose the longitudinal sequences in Figure 1 represents the observed day-wise visiting sequences of individuals as detected by (inaccurate) sensor(s). Black shade represents presence of an individual during the respective time interval deterministically. Unshaded region represents a non-deterministic value or uncertainty about the individual's presence or absence.

Temporal data collected through sensors can have uncertainty as individual may actually be absent during the time interval or the sensor may have failed to detect the presence of an individual (false negative). This may lead to an ambiguity in the observed data. Another aspect of the behavior data is temporal variability of underlying patterns because human routine behavior is not perfectly periodic. Moreover, human behavior is multi-modal in nature and the number of these modes are unknown beforehand. In the example, individuals exhibit four modes of mobility behavior where cluster A, B, and C represent patterns spanning entire time period whereas cluster D represents a localized pattern. The time of occurrence and duration of \textit{localized pattern} is also unknown which poses another challenge.

In addition, to determine number of modes of behavior ($C$) in a person‘s visiting sequence is not an easy task. The optimal choice of \textit{C} strikes a balance between maximum compression of the data using a single cluster, and maximum accuracy by assigning each data point to its own cluster. There are many techniques, like \textit{The Elbow Method} or \textit{Silhouette index} to estimate number of clusters. These techniques do not work well in non-euclidean space. One of such technique involves $PCA$ analysis before clustering and setting \textit{C} equal to the number of dominant principal components. This method is not accurate because it often underestimates number of clusters in the dataset. 

Furthermore, the choice of distance measure is of utmost importance for any clustering. The conventional distance measures, like euclidean, Manhattan, Jaccard, etc., compare corresponding time slots and do not capture temporal dynamics or affinity between neighbouring time instances which is very important in case of uncertainty, noise and variability in data. Moreover, they are not sensitive to local differences between sequences.

To handle the aforementioned issues, in this paper, we focus on extracting and identifying visiting (or, mobility) behavior of individuals to a location (named, MTpattern) which requires to group or cluster similar visiting sequences. We  propose a novel distance measure scheme to find an appropriate dissimilarity measure between visiting sequences that is invariant to only small temporal variation and uncertainty in the observation. Then, we use segment tree data structure to discover localized frequent patterns in given time window or to find frequent patterns of given length efficiently. Then, we propose an effective clustering mechanism by exploiting affinity propagation to cluster sequences with the given constraints and unknown number of modes (or clusters) of behavior. We consider the average of all visiting sequences in a cluster for every unit time interval to have a \emph{behavior pattern} of individuals. 

We formulate our problem of finding behavior patterns problem as a \emph{constrained optimization problem} where we minimize number of mutually exclusive clusters that cover the entire dataset under the given constraint of maximum permissible local dissimilarity enforced by dissimilarity metric\footnote{This problem is similar to finding minimum \emph{dominating set} of vertices in an undirected graph}. We perform experiments on a simulated and two real-world datasets. Results show that \textit{MTpattern} outperforms three baseline clustering algorithms.

The contributions of this paper are as follows:
\begin{enumerate}

\item We propose a novel approach to discover visiting  (or,  mobility) behavior patterns of  individuals of given length by clustering similar visiting sequences or to find localized frequent patterns in a given time window (named, MTpattern).

\item We represent time series data into discretized sequence of regularly sampled equal sized unit time intervals for better analysis and propose a novel dissimilarity measure, called TDist, to cluster similar sequences by putting an upper-bound on the local dissimilarity for handling temporal variations and uncertainty in the data. Then, we fine-tune a non-parametric exemplar based clustering  technique to cluster  sequences  with  the  given  constraints  and  unknown number  of modes of behavior.

\item  To validate and to show the usability of \textit{MTpattern}, we evaluate our proposed approach on two real-world datasets and a simulated dataset. Results show that \textit{MTpattern} outperforms three baseline clustering algorithms w.r.to both internal and external measures. 
\end{enumerate}

\section{Related Work}
\label{sec:related_work}
The idea of discovering underlying patterns in individuals' behavior is not new. In this section, we shall discuss important and popular techniques for modeling mobility behavior using time-series data. Majority of the works have employed time-series clustering algorithms to infer behavior patterns which can be grouped into two main categories: \textit{partitional and hierarchical clustering}, and \textit{model-based clustering}. We also discuss \textit{sequential pattern mining} and \textit{PCA-based Eigenbehavior techniques} that are commonly used to find the structure of frequent individuals' behavior. We also discuss the limitations and shortcomings of these existing techniques.

\subsection{Time-series Clustering for Human Behavior Modeling}

\subsubsection{Partitional and Hierarchical Clustering}
Previous works have used partitional (like, K-Means or K-Medoids) \cite{WarrenLiao20051857, t2017} or hierarchical \cite{ldh10} clustering techniques for behavior modelling. For these techniques, user has to either pre-specify the number of clusters or set an upper bound on overall representational error to stop clustering. Determining number of clusters in advance or maximum permissible representational error is not trivial and is often subjective.



\subsubsection{Model Based Clustering}
Under \emph{model based clustering} (HMM, Kalman filter, etc.) probabilistic models are initially built on time-series. The asymmetric dissimilarity between time-series $A$ from $B$ is calculated by using a posteriori probability of time-series $A$ given the probabilistic model of time-series $B$. Probabilistic distance measures, such as KL distance \cite{Bicego2003,Zhong,ICML2011Li159,KlForHMM} are popularly used. The major limitation of this approach is that there is need to pre-determine number of clusters. Matsubara, et al. \cite{matsubara2014autoplait} used an Baum-Welch algorithm based approach for the finding distinct high-level patterns from a large set of co-evolving sequences. The Baum–Welch algorithm is a special case of the EM algorithm used to find the unknown parameters of a HMM. However, model based sequence clustering methods, such as HMM are more sensitive to the order of events and invariant to the actual time of occurrence of the events. In model based clustering, it is also important to assume that the observations are deterministic which is not possible in the real-life data as data often contains uncertainty and non-deterministic instances in observation.

\subsection{Sequential Pattern Mining}
In recent times, priori-based GSP algorithm \cite{Srikant1996}, projection-database based Freespan \cite{1339268} and Prefix Span \cite{Han:2000:FFP:347090.347167}, vertical id-list database and lattice theory based SPADE \cite{Zaki2001} algorithm has been used to find frequent subsequences in a given set of categorical time sequences. These methods only consider the topological order of events for finding frequent patterns instead of the absolute time of occurrence or duration of the events. Moreover, in case of interval events unlike instantaneous events, it is not trivial to establish the topological order. Some recent works \cite{4733940,5360529,BenZakour2012} have proposed mining temporal interval patterns from interval data. For \emph{extracting temporal patterns from interval-based sequences} \cite{Guyet:2011:ETP:2283516.2283614}, Guyet, et al. \cite{Allen:1983:MKT:182.358434} proposed converting temporal sequences into sequence of events linked by Allen's temporal relations. Rawassizadeh, et al. \cite{multivariate2016} proposed a model for identifying frequent behavioral patterns with temporal granularity from the real-time dataset. But they dealt with temporal events that fit into the Allen’s interval algebra, which is not about time-series analysis. Even the dissimilarity measure (City-Block distance) in \emph{extracting temporal patterns from interval-based sequences} between subsequences does not take into account uncertain (or, undeterministic) observations in data.

\subsection{PCA Decomposition}
Another popular method to model time-series data is `Eigenbehavior' \cite{NathanEigStructure}. These eigenbehaviors are the eigenvectors of the covariance matrix of behavior data obtained after application of PCA or Singular Value Decomposition (SVD). The first few eigenvectors (highest eigenvalues) of the decomposition typically account for a very large percentage of the overall variance in the longitudinal data's eigen decomposition. It is claimed that every non-trivial (high eigenvalue) eigenvector represents a recurrent dominant pattern. One advantage of this approach is that we do not need to pre-determine or predict the number of modes in the data and it can be directly derived from the number of eigenvectors having large eigenvalues. But, it still suffers from a number of shortcomings. Like conventional distance measures, it fails to capture the temporal dynamics as it plots the sequence in Euclidean space and treats every dimension independent of each other failing to capture the affinity between nearby time instances. Number of eigenvectors is limited by number of dimensions and if a cluster can be represented by a linear combination of eigenvectors having higher eigenvalue, then a new eigenvector in the direction of the cluster is not needed. Therefore, PCA often underestimates number of clusters. Since, all the eigenvectors need to be orthogonal to each other, many eigenvectors do not point in the direction of actual clusters since not all clusters are orthogonal to the principal eigenvector at mean. So, all cluster centroids may not necessarily lie on some eigenvector. Eigenvectors are also biased towards cluster located at a larger distance from mean as they cause bigger variance in the data.

Our proposed approach handles the issue of temporal dynamics, uncertainty in the data, and do not require  number of behavior modes as input beforehand at the same time. To the best of our knowledge, no prior work studies human behavior patterns mining by handling the above-mentioned issues simultaneously. 

\begin{table}[h!]
\begin{tabular}{ |p{1.1cm}||p{7cm}| }
 \hline
 Notations & Description \\
 \hline
 $m$ &	Number of individuals. \\
 \hline
$L$ &	Total discretized timestamps in a BIS or length of the BIS in a day.  \\
\hline
$e$ &	Number of days. \\
\hline
$d(A, B)$ &	Distance from one interval sequence $A$ to another interval sequence $B$. \\
\hline
$D(A, B)$ &	Distance from one interval sequence $A$ to another interval sequence $B$ and vice versa. \\
\hline
$\delta$ &	A threshold to determine two points in a time point sequence will be in the same time interval in time interval sequence or not. \\
\hline
$\lambda$ &	It is the length of equal sized unit intervals in the discretized time interval sequence (BIS). \\
\hline
$\Omega$ &	It is a threshold to handle uncertainty in observed data and temporal variability in patterns. \\
\hline
C &	Number of clusters. \\
\hline
cnt	& Number of discrete intervals in a BIS. \\
\hline
$t\textsubscript{As}$ & Timestamp at the starting of a time interval in a BIS $A$. \\
\hline
$t\textsubscript{Ae}$ & Timestamp at the ending of a time interval in a BIS $A$. \\
\hline
ITDist & An interval temporal distance between any two time intervals $A$ = [$t\textsubscript{As}$, $t\textsubscript{Ae}$] and $B$ = [$t\textsubscript{Bs}$, $t\textsubscript{Be}$], denoted as $ITDist(A, B)$ which is the absolute difference of the means of two time intervals, given by 
$|\frac{(t\textsubscript{As} + t\textsubscript{Ae})}{2} - \frac{(t\textsubscript{Bs} + t\textsubscript{Be})}{2}|$.  \\
 \hline
\end{tabular}
\caption{Notations and their descriptions}
\label{table:1}
\end{table}

\section{Definitions and Problem Statement}
Human behavior at a location can help us to infer interesting information about the location. Simple statistics like number of visitors, average stay time and frequency can reveal semantics of a location. There have been a number of applications in a
number of domains, such as transportation (analyze the number of visitors to identify time of the day when overcrowding usually takes place \cite{liu2006scalable}), smart environment (in a residential location, an intelligent lighting, heating or cooling system can be developed by modelling the presence of people in the room \cite{pan2013trace}\cite{chakravorty2013privacy}), health (monitoring the habits and mobility of patients can be used as an indicator of overall health \cite{soulas2013monitoring}), education (understanding how the campus is used can provide very important information and insights to college authorities \cite{wang2014studentlife}), etc. We define the problem as follows: 

\textbf{Problem 1.} Given a raw sensor data tracking the presence of individuals visiting a location over a long period of time and a threshold $\delta$, assuming that presence of individuals in a location is not strictly periodic, sensors are not accurate and may produce false negatives, and multi-modality is inherent in the individuals' visiting patterns and the number of these modes are unknown beforehand, the goal is to transform the raw sensor data into discretized sequence of regularly sampled equal sized unit time intervals for better analysis and then effectively summarize the frequent behavioural patterns of \textit{m} individuals in any given time window during the day.

To be able to formulate the problem first, we describe our definitions. Table 1 lists notations that we have used in this paper.

Human behavior is recurring and influenced by a range of factors (such as, time). Here, human behavior under the influence of time has been called "frequent behavioral patterns".

\textbf{Definition 1.} Time Point Sequence (PS) is an ordered sequence of \textit{p} timestamps at a day \textit{d} at which an individual \textit{i} was detected by the sensor(s), 
$PS^d_i$ = $\langle$\{$t\textsubscript{1}$\}, \{$t\textsubscript{2}$\},...,\{$t\textsubscript{p}$\} $\rangle$ \, \textrm{for} \, 1 $\leq$ i $\leq$ m \, \textrm{where} \, \textit{m} \, \textrm{is the number of individuals}.

\textbf{Definition 2.} Time Interval Sequence (IS) is an ordered sequence of \textit{q} time intervals at a day $d$ computed from the \textit{PS}, $IS_i^d$ = $\langle$\{$t\textsubscript{1s}$, $t\textsubscript{1e}$\}, \{$t\textsubscript{2s}$, $t\textsubscript{2e}$\},..., \{$t\textsubscript{qs}$, $t\textsubscript{qe}$\} $\rangle$ where $t\textsubscript{1s}$ and $t\textsubscript{1e}$ represents the start and end of interval, and 1 $\leq$ i $\leq$ m where $m$ is the number of individuals.

\textbf{Definition 3.} Discretized Time Interval Sequence (BIS) is a sequence of regularly sampled equal sized unit intervals at a day $d$ computed from the IS, $BIS_i^d$ = $\langle$\{0, $\lambda$\}, \{$\lambda$, 2*$\lambda$\},..., \{($L$-1) * $\lambda$, $L$ * $\lambda$\}$\rangle$  where $\lambda$ represents length of unit interval along with the binary value (0 or 1) representing whether the individual was detected or not by the sensor in the corresponding time interval, \textit{L} is the total length of BIS in a day, and 1 $\leq$ i $\leq$ m where $m$ is the number of individuals. We will talk more on how we set value of $\lambda$ in the Section 6.1.3. In addition, transforming raw sensor data into Discretized Time Interval Sequence (BIS) is discussed in Section 4.1. A Discretized Time Interval Sequence (BIS) is also called visiting sequence of an individual.

\textbf{Definition 4.} Partial $\Omega$ covering for BIS $A$ is set of all other BISs $B\textsubscript{i}$, 1 $\leq$ i $\leq$ m such that for every unit time interval with value 1 in $A$, there exists some time interval with value 1 in $B\textsubscript{i}$ under the given constraint of maximum permissible  local  dissimilarity, i.e.,  ITDist($A$, $B\textsubscript{i}$) < $\Omega$.

\textbf{Definition 5.} Complete $\Omega$ covering for BIS $A$ is set of all other BISs $B\textsubscript{i}$, 1 $\leq$ i $\leq$ m, such that $B\textsubscript{i}$ is in \emph{partial $\Omega$ covering} of $A$ and $A$ is in \emph{partial $\Omega$ covering} of $B\textsubscript{i}$ at the same time.

\textbf{Definition 6.} Frequent behavioral patterns are the average of all BISs in a cluster, such that every member of the cluster is in complete $\Omega$-covering of an \emph{exemplar} BIS which is also a member of the cluster. The average obtained represents the probability of the presence of  individuals in every unit time interval for that particular pattern.

\begin{figure}[htbp]
\centerline{\includegraphics[scale= 1.3]{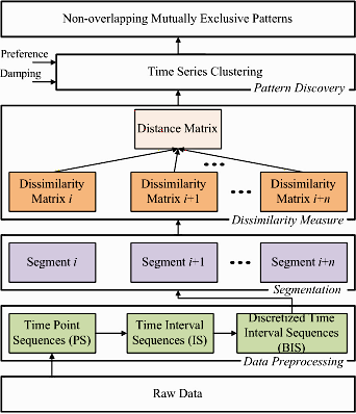}}
\caption{\textit{MTpattern} Overview}
\label{fig:framework}
\end{figure}

\section{Our Proposed Solution}
In this section, we discuss about our proposed solution, \textit{MTpattern} in detail.


Figure \ref{fig:framework} shows the overview of our proposed approach \textit{MTpattern}. \textit{MTpattern} is composed of four major parts as follows: 1) \textit{Data Preprocessing.} We calculate the \textit{Time Interval Sequence} (IS) for every individual from its corresponding \textit{Time Point Sequence} (PS). Then the \textit{Time Interval Sequences} are discretized into sequence of regularly sampled equal sized unit intervals, named \textit{Discretized Time Interval Sequence} (BIS) for better analysis.  2) \textit{Segmentation.} To facilitate the piece-wise analysis, each and every BIS is hierarchically segmented and stored in a segment tree. 3) \textit{Dissimilarity Measure.} To calculate symmetric dissimilarity between pair of same length BIS segments that is invariant to uncertainty in observations and small temporal variation in underlying patterns, we propose a novel symmetric dissimilarity metric, called TDist. The  dissimilarity matrix is pre-computed for every segment in the segment tree. Dissimilarity matrix (or distance matrix) can then be computed for any time interval from dissimilarity matrix of segments in the segment tree. 4) \textit{Pattern Discovery}. Every row of the  dissimilarity matrix represents a cluster which is the complete $\Omega$-Covering of the corresponding BIS. These clusters are overlapping in nature as any BIS may be a member of more than one complete $\Omega$-Covering. Therefore, we further extend our analysis and optimize the discovery of patterns by finding a minimum set of disjoint (or, non-overlapping) clusters such that every BIS is a member of exactly one cluster and every cluster has an exemplar BIS that has all members of the corresponding cluster in it is complete $\Omega$-Covering. 

\begin{figure*}[h!]
  \centering
  \begin{subfigure}[t]{0.3\textwidth}
    \centering
    \includegraphics[scale = 0.45]{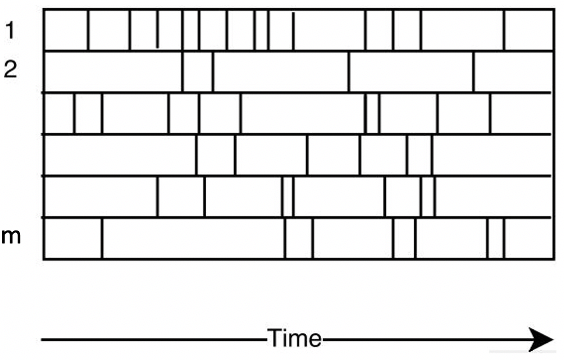}
    \caption{\label{fig:fig1} \textit{Time Point Sequence (PS) of m individuals}}
  \end{subfigure}
  ~
  \centering
  \begin{subfigure}[t]{0.3\textwidth}
    \centering
    \includegraphics[scale = 0.3]{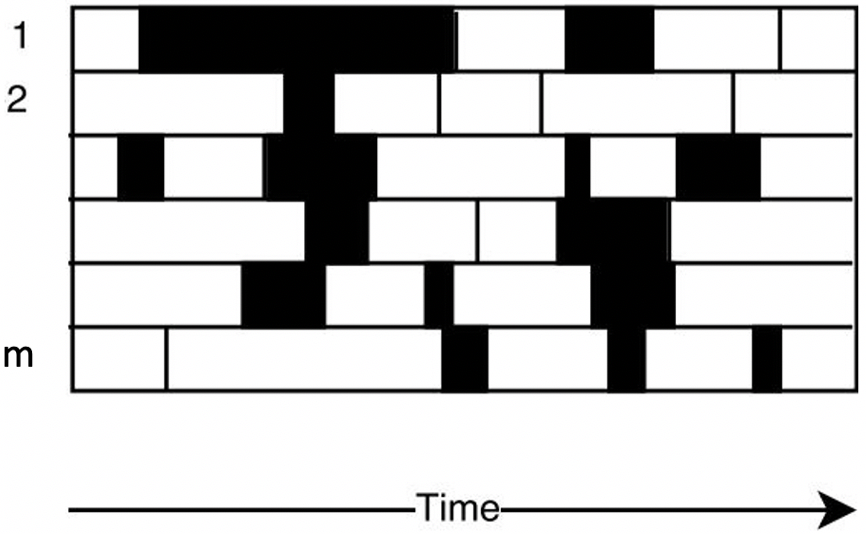}
    \caption{\label{fig:fig2} \textit{Time Interval Sequence (IS) of m individuals}}
  \end{subfigure}
  ~
  \centering
  \begin{subfigure}[t]{0.3\textwidth}
    \centering
    \includegraphics[scale = 0.48]{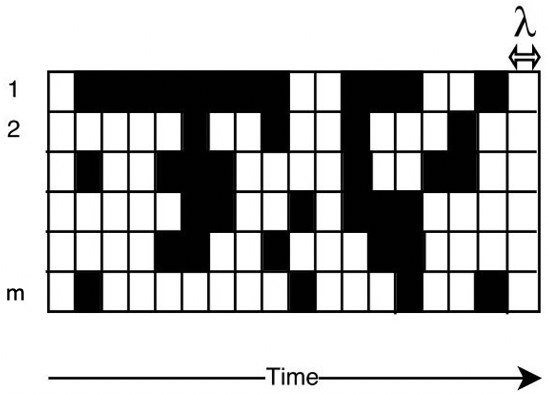}
    \caption{\label{fig:fig3} \textit{Discretized Time Interval Sequence (BIS) of m individuals}}
  \end{subfigure}

 \caption{Preprocessing of raw sensor data}
 \label{fig:preProcessing22}
\end{figure*}

To minimize the number of clusters, the dissimilarity matrix of the time window is fed into affinity propagation module and it's \emph{preference} parameter tuned to minimize number of unique $\Omega$-coverings that cover all the BISs while maximizing the net similarity between member BISs of the cluster and the exemplar BIS of the cluster it is part of. We take the average of all BIS in $\Omega$-covering which is a discrete probability distribution as a representative of the corresponding cluster.

\subsection{Data Preprocessing}
We analyze and preprocess the collected raw sensors data for better analysis. Figure \ref{fig:preProcessing22} (a) shows the Time Point Sequence (PS) for $m$ individuals in which a vertical bar ($|$) in a row shows the instantaneous timestamp when a WiFi packet corresponding to a particular MAC-address is received (i.e., an individual presence is detected). However, storing information of closely spaced presence is redundant and costly. Therefore, we calculate the \emph{IS} for every individual from its corresponding \emph{PS} by inspecting the time delay between consecutive individual's presence detected in $PS$. If the time delay between consecutive bars ($|$) (i.e., consecutive WiFi packets captured of an individual) is below a threshold $\delta$, then they form a time interval in the \emph{IS}. We discuss in detail on how appropriate interval threshold $\delta$ is calculated in Section 6.1.3. Figure \ref{fig:preProcessing22} (b) shows $m$ \emph{IS} where each filled rectangle (\rule{0.3cm}{0.2cm}) shows the time interval when an individual is present. However, performing piece-wise analysis on \emph{IS} is difficult as the intervals in continuous time domain are of varying length (or duration). So, the \emph{IS} are discretized into sequence of regularly sampled equal sized unit intervals of length $\lambda$. We set $\lambda$ equal to $\frac{\delta}{2}$ so that any gap in raw sensor data (consisting of point sequences) which is more than $\delta$ minutes is captured in BIS after discretization. We obtain BIS after representing \emph{IS} in discrete time domain. `1'  represents a deterministic value and `0' represents a non-deterministic value. Figure \ref{fig:preProcessing22} (c) shows discretized time interval sequence (BIS) of $m$ individuals. 

\subsection{Segmentation}
\label{sec:segmentation}
\begin{figure}[h!]
\centering
\includegraphics[height=6cm , width =9cm]{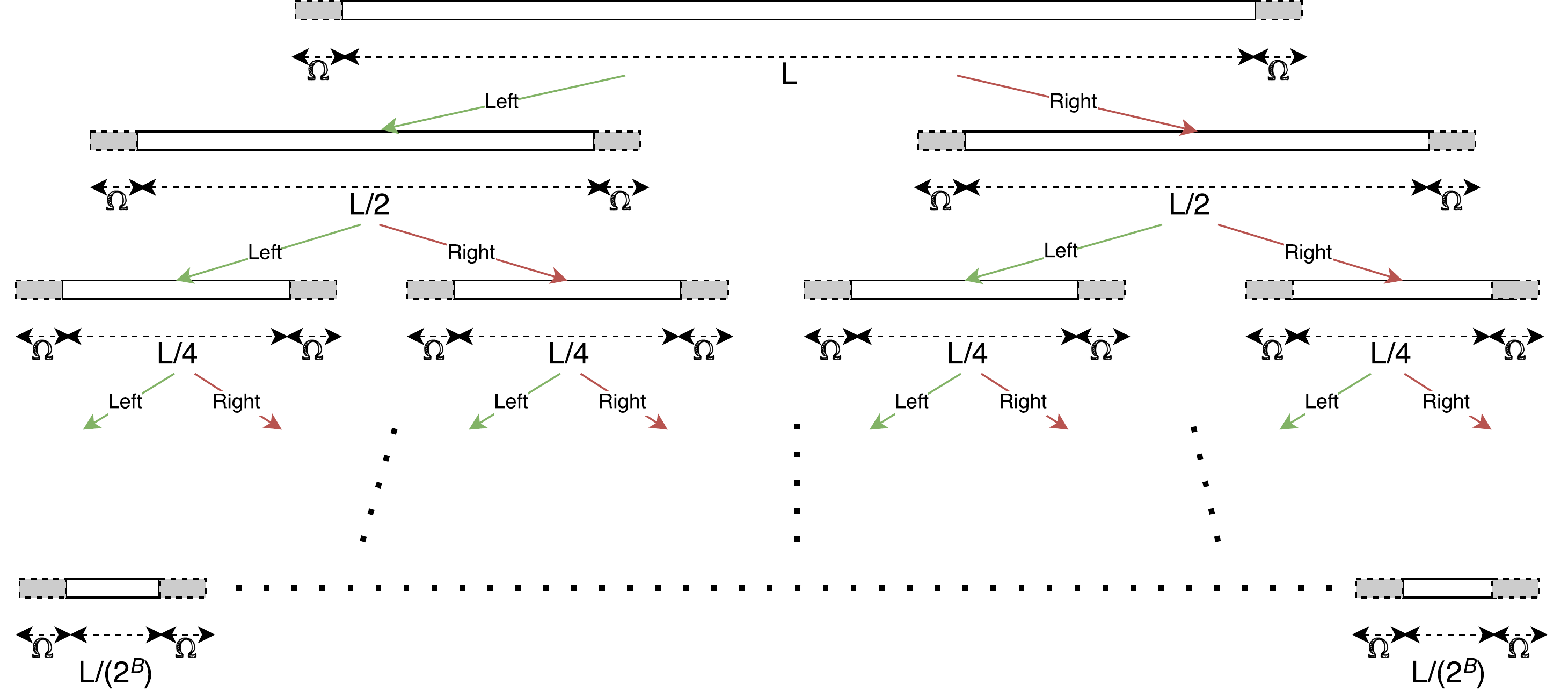}
\caption{Hierarchical segmentation of a BIS}
\label{fig:segmentation}
\end{figure}

There can be a possibility that a pattern spans for an entire time period (a \emph{day} in our case) or only during some time window during the day. Since, we do not know when and for how long this regular behavior occurs during the day, we segment every BIS into hierarchical segments to facilitate piece-wise analysis. These segments are arranged in a binary tree data structure called \textit{segment tree} as shown in Figure \ref{fig:segmentation}. Every segment is divided into two equal parts in the next level in the segment tree hierarchy. The $Left$ child of any segment in the tree is the left half of the segment and the $Right$ child is the right half off the segment. This data structure is particularly helpful when a solution to a problem can be represented as a combination of solutions to it's sub problems. This is true in our work as any frequent behavior pattern is also piece-wise frequent patterns and can be represented as a concatenation of frequent patterns in it's smaller pieces or segments.

The length of the segments at the leaf nodes of the segmented tree represents the highest level of granularity $\frac{L}{2^B}$ where $B$ is a positive integer in (0, ${\log_2 L}$) range. $B = 0$ corresponds to lowest granularity, whereas $B =\log_2 L$ represents highest granularity possible (as the segments at the leaf nodes will be of unit length). Any time interval in continuous $[start, end]$ can be approximated to discretized time domain. Higher granularity will give more accurate results but increase the time and space complexity of the algorithm and makes it more computationally intensive. So, there is need of a trade-off between accuracy and complexity. Therefore, we set $\frac{L}{2^B}$ equal to ${\Omega}$.

Every BIS segment is also augmented with extensions at both ends. These extensions are used for computing partial dissimilarity between segments at borders and ensure that no information is lost near the segment boundaries due to segmentation. These extensions only contain $\frac{\Omega}{\lambda}$ discretized time internal units because when we compute dissimilarity between two BIS segments for every unit time interval with value `1' in one BIS, we only need to inspect unit time intervals in the other BIS with interval temporal distance less than $\Omega$ (discussed in sub-section \ref{sec:dissimilarity}). The augmented BIS with extensions is represented by \emph{eBIS}.

\subsection{Dissimilarity Measure}
\label{sec:dissimilarity}

In our work, every BIS is a data point in non-Euclidean space. We define dissimilarity metric between pair of BISs by comparing the relative time of occurrence of unit intervals with value `1' in the two BIS. For every unit interval with value `1' in BIS\textsubscript{1}, we calculate $ITDist$ of nearest unit interval with value `1' in BIS\textsubscript{2}. The average of $ITDist$ value for all unit time intervals with value `1' in BIS\textsubscript{1} from the nearest unit time interval in BIS\textsubscript{2} gives the partial dissimilarity of BIS\textsubscript{1} from BIS\textsubscript{2} denoted by d(1, 2). We also record the count of unit time intervals with value `1' in BIS\textsubscript{1} as $cnt_A$. We repeat the process to find partial dissimilarity of BIS\textsubscript{2} from BIS\textsubscript{1} given by d(2, 1) and record the number of unit time intervals with value `1' in BIS\textsubscript{2} as $cnt_B$ (see Algorithm 1 and 2).

\begin{figure}[htbp]
  \centering
  \begin{subfigure}{1\linewidth}
    \centering
    \includegraphics[scale=0.4]{fig/DistanceMeasureWithExtensionAB12.png}
    \caption{\label{fig:fig1}$B$ is in partial $\Omega$ covering of $A$}
  \end{subfigure}   
    \centering
  \begin{subfigure}[b]{1\linewidth}
    \centering
    \includegraphics[scale = 0.4]{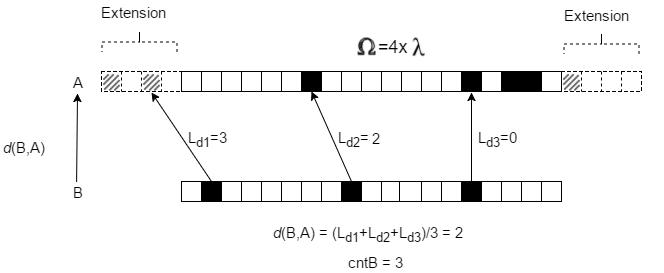}
    \caption{\label{fig:fig2}$A$ is in partial $\Omega$ covering of $B$}
  \end{subfigure}
 \caption{\textbf{Distance between $A$ and $B$ is within bounds} - 
 When the time difference for every unit interval with value `1' from the nearest unit time interval in the other BIS is \textit{less than} $\Omega$ then the complete dissimilarity between the pair of BIS, given by $D$($A$, $B$) = (\begin{math}\frac{d(A, B) * cnt_A + d(B, A) * cnt_B}{cnt_A + cnt_B} = 1.71 \end{math}) is defined}
 \label{fig:defined_distance}
\end{figure}

\begin{figure}[htbp]
  \centering
  \begin{subfigure}{1\linewidth}
    \centering
    \includegraphics[scale=0.4]{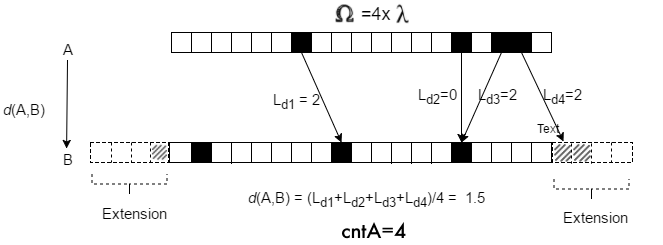}
    \caption{\label{fig:fig1}$B$ is in partial $\Omega$ covering of $A$}
  \end{subfigure}   
  \centering
  \begin{subfigure}[b]{1\linewidth}
    \centering
    \includegraphics[scale=0.4]{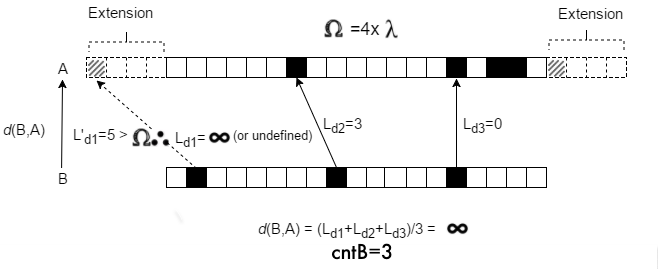}
    \caption{\label{fig:fig2}A is \textbf{NOT} in partial $\Omega$ covering of B}
  \end{subfigure}
 \caption{\textbf{Distance between $A$ and $B$ is out of bounds} -
 When the interval temporal distance of any unit interval with value `1' from the nearest unit time interval in the other BIS is \textit{more than} $\Omega$ then the complete or overall dissimilarity between the pair of BIS, given by $D$($A$, $B$) is undefined (or, infinite)}
 \label{fig:undefined_distance}
\end{figure}

\begin{algorithm}[h!]
\caption{Find nearest interval of BIS1 from index i}\label{min_itd}
\begin{algorithmic}[1]
\Procedure{MIN\_ITDist}{i, BIS1, $\Omega$}
\For {\text{\textit{index} in range(0, $\Omega$)}}
\If{BIS1(\textit{i} + \textit{index}) = 1 or BIS1(\textit{i} - \textit{index}) = 1}
\State \textbf{return} \textit{index}
\EndIf
\EndFor
\EndProcedure
\State \textbf{return} $\infty$
\end{algorithmic}
\end{algorithm}

\begin{algorithm}[h!]
\caption{Partial Distance from BIS1 to BIS2}\label{partial_dist}
\begin{algorithmic}[1]
\INPUT  
\Statex $BIS1$, \Comment{A Discretized Time Interval Sequence}
\Statex $BIS2$, \Comment{A Discretized Time Interval Sequence}
\Statex $\Omega$, \Comment {Threshold for local distance}
\OUTPUT
\Statex $d(BIS1, BIS2)$, \Comment{Partial Distance from BIS1 to BIS2}
\Statex $cnt$, \Comment{number of discrete intervals (or '1') in BIS1}
\Procedure{$P\_D$}{\textit{BIS1}, \textit{BIS2}, $\Omega$} \Comment{partial distance from BIS1 to BIS2}
\State $\textit{Len} \gets \text{length of }\textit{BIS1} = \text{length of }\textit{BIS2}$
\State $d(\textit{BIS1}, \textit{BIS2}) = 0$
\State $cnt \gets 0$
\For{\text{\textit{i} in  range(0, \textit{Len})}}
\If {\textit{BIS1(i)} = 1}
\Statex \parbox[t]{3in}{\raggedleft /*eBIS2.intervals is set of discretized intervals in extended BIS2*/}
\State $L\textsubscript{di} \gets MIN\_ITDist(i, eBIS2, \Omega)$
\If {\text{\textit{L}\textsubscript{di} = $\infty$}}
\State \textbf{return} $\infty$ 
\EndIf
\State $d(\textit{BIS1}, \textit{BIS2})\gets d(\textit{BIS1},\textit{BIS2}) + L\textsubscript{di}$
\State $cnt \gets cnt + 1$
\EndIf
\EndFor
\State\textbf{return} $d(\textit{BIS1},\textit{BIS2}), cnt$ 
\EndProcedure
\end{algorithmic}
\end{algorithm}

\begin{algorithm}[ht!]
\caption{Proposed Distance Measure, TDist}\label{d_msr}
\begin{algorithmic}[1]
\INPUT  
\Statex $BIS1$, \Comment{A Discretized Time Interval Sequence}
\Statex $BIS2$, \Comment{A Discretized Time Interval Sequence}
\Statex $\Omega$, \Comment {Threshold for local distance}
\OUTPUT
\Statex $D(A,B)$, \Comment{Distance between BIS1 and BIS2}
\Procedure{Complete\_Dist}{\textit{BIS1}, \textit{BIS2}, $\Omega$}
\State $d(BIS1, BIS2), cnt1 \gets P\_D(\textit{BIS1}, \textit{BIS2}, \Omega)$
\State $d(BIS2, BIS1), cnt2 \gets P\_D(\textit{BIS2}, \textit{BIS1}, \Omega)$
\State $D(A,B) \gets ( d(BIS1, BIS2) + d(BIS2, BIS1) ) / (cnt1 + cnt2)$
\State \textbf{return} $D(A,B)$
\EndProcedure
\end{algorithmic}
\end{algorithm}

\begin{figure}[htbp]
\centering
\includegraphics[scale = 0.35]{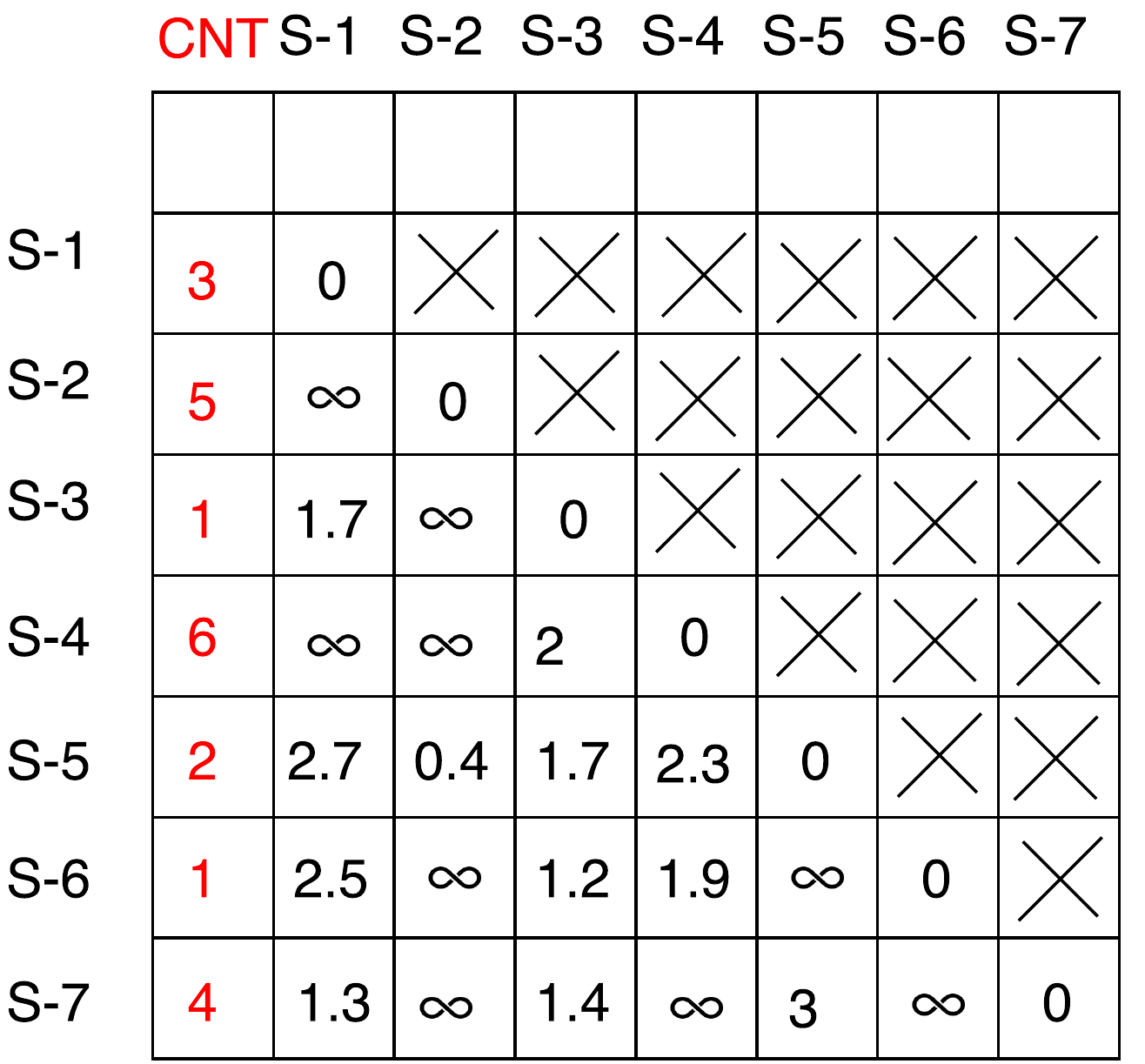}
\caption{Sample Distance Matrix for 7 BISs}
\label{fig:dissimilarity_matrix}
\end{figure}

Figure \ref{fig:defined_distance} and \ref{fig:undefined_distance} illustrate the dissimilarity measure computation with two examples. If $ITDist$ for any unit time interval with value `1' in either of the two BIS from the nearest unit time interval with value `1' in the other BIS is greater than some threshold $\Omega$ then the partial dissimilarity computation algorithm is halted and the overall dissimilarity between the two BIS is set as infinite (or, undefined) which means that the two BISs can never be linked together (or, one of the two BISs can never be exemplar of the other BIS).  The overall dissimilarity between BIS\textsubscript{1} and BIS\textsubscript{2} ($D$(1, 2) = $D$(2, 1)) is calculated from the partial dissimilarities given by Algorithm 2. The dissimilarity matrix / distance matrix thus obtained is \textit{symmetric} about main diagonal, so only one half needs to be stored (see Figure \ref{fig:dissimilarity_matrix}).  


The value of $\Omega$ takes into account uncertainty in observed data and temporal variability of patterns. It has been proved that temporal patterns of human behavior tend to be normally distributed \cite{5f5537a223c84269ba2508ba6845cde5}. This property can be used to model starting (arrival) and ending (departure) time of behavior patterns without uncertainty as normal distributions. But the \emph{observed} behavior patterns contain uncertainty (or, false negatives). The distribution with the combined effects of normal temporal behavior of humans and uncertainty in observations manifests itself in the \emph{observed} arrival and departure distribution of every behavior pattern. To get a sample of this distribution, we record the timestamp (in discrete time domain) when a particular individual was first detected during the day. 

We construct dissimilarity matrix / distance matrix for any time interval from dissimilarity matrix of it's constituent segments contained in a segment tree. Let the complete dissimilarity between two BISs $A$ and $B$ of same length be $D$($A, B$) (see Algorithm 3). We divide $A$ and $B$ into $k$ segments. Let the segment $i$  of $A$ and $B$ is $A\textsubscript{i}$ and $B\textsubscript{i}$, respectively, and the complete dissimilarity between them is $D$($A\textsubscript{i}$, $B\textsubscript{i}$). Let $Cnt(A\textsubscript{i})$ and $Cnt(B\textsubscript{i})$ be number of unit time intervals with value `1' in the BIS $A\textsubscript{i}$ and BIS $B\textsubscript{i}$, respectively as described in Section \ref{sec:dissimilarity}. Then, $D$($A, B$) can be written as 

\begin{equation}
D(A, B) =  \frac{\sum_{i=0}^{k} D(A\textsubscript{i}, B\textsubscript{i})\times (Cnt(A\textsubscript{i}) + Cnt(B\textsubscript{i}))} {\sum_{i=0}^{k} Cnt(A\textsubscript{i}) + Cnt(B\textsubscript{i})}
\label{eq:segment_dissimilarity}
\end{equation}

Using the Eq. \ref{eq:segment_dissimilarity}, \textit{MTpattern}  calculates the dissimilarity between every pair of BIS in any time interval from the pair-wise dissimilarity score of its constituent BIS segments in the segment tree. So, \textit{MTpattern} can reconstruct dissimilarity matrix for any time interval without the need to compute dissimilarity score between every pair of BIS from scratch. Algorithm 4 shows how we can query the segment tree and retrieve the dissimilarity matrix of all the constituent segments of a time interval ([Le, Ri]). If the length of the time interval in a BIS (number of unit time intervals) is $L$ then the segment tree returns ${O}(\log_{2}{L})$ segments in the worst case.

\begin{algorithm}
\label{algo:combine}
\caption{Retrieve list of Distance Matrix of constituent segments of an interval}\label{combine}
\begin{algorithmic}[1]
\INPUT
\Statex $[Le,Ri]$ , \Comment {Query Interval endpoints}
\Statex $root$ , \parbox[t]{3in} {\Comment {\raggedleft root of the segment\_tree. Each node in the tree contains end points and distance matrix for the corresponding segment}}
\OUTPUT
\Statex list(distance\_matrix), \parbox[t]{2.05in}{\Comment{\raggedleft List of distance matrices for segments of interval $[Le, Ri]$}}
\Procedure{Get\_D\_Matrix}{$root$, $[Le,Ri]$}
\If {$root.left>=Le$ and $root.right<=Ri$}
\State \textit{return} $root.dist\_matrix$
\ElsIf{$root.left>=Ri$ or $root.right<=Le$}
\State \textit{return null}
\Else
\State $lft\_list$ $\gets$  Get\_D\_Matrix$(root.left$, $[Le,Ri])$
\State $rght\_list$ $\gets$  Get\_D\_Matrix$(root.right$, $[Le,Ri])$
\State \textit{return} $[lft\_list,rght\_list]$
\EndIf
\EndProcedure
\end{algorithmic}
\end{algorithm}

Distance measure between any two BIS ensures that for every interval in a sequence member of $\Omega$-covering, there exists an interval in exemplar BIS such that their ITDist is less than a given threshold $\Omega$. We also call this minimum ITDist between constituent intervals as local distance. Transitivity property ensures that maximum local distance between any two BISs in a $\Omega$-covering is less than 2 * $\Omega$.

\subsection{Pattern Discovery}
\label{sec:pattern_discovery}
In this section, we shall discuss discovery of behavior patterns from the database of BISs.

The dissimilarity matrix represents overlapping BIS clusters where every row represents a complete $\Omega$-covering of the corresponding BIS (see Figure \ref{fig:dissimilarity_matrix} for example). These $\Omega$-coverings are overlapping as any given BIS may be a member of complete $\Omega$-covering of more than one BIS. If cardinality of a complete $\Omega$-covering of a BIS is small, the corresponding behavior pattern is not frequent. $\Omega$-coverings which are a subset of other $\Omega$-coverings can also be ignored. If two $\Omega$-coverings are equal then \textit{MTpattern} will ignore the $\Omega$-covering for which average distance of all BIS with the exemplar BIS of the corresponding $\Omega$-covering is higher. If the cardinality is more than a threshold $\alpha$ then \textit{MTpattern} takes the average of all BIS in a $\Omega$-covering for every unit time interval to obtain a behavior pattern in the form of discrete time probability distribution.
For example, in Figure \ref{fig:dissimilarity_matrix}, if the threshold $\alpha$ is 3, then,

\begin{itemize}
    \item $\Omega$-covering of S-1 is [S-1, S-3, S-5, S-6, S-7] and is frequent
    \item $\Omega$-covering of S-2 is [S-2, S-5] and is not frequent
    \item $\Omega$-covering of S-3 is [S-1, S-3, S-4, S-5, S-6, S-7] and is frequent
    \item $\Omega$-covering of S-4 is [S-3, S-4, S-5, S-6] and is frequent
    \item $\Omega$-covering of S-5 is [S-1, S-2, S-3, S-4, S-5, S-7] and is frequent
    \item $\Omega$-covering of S-6 is [S-1, S-3, S-4, S-6] and is frequent
    \item $\Omega$-covering of S-7 is [S-1, S-3, S-5, S-7] and is frequent
\end{itemize}

Since we are not concerned about the $\Omega$-coverings which are subset of other $\Omega$-coverings, we ignore $\Omega$-coverings of S-2, S-4 and S-7.

\subsubsection{Optimization}
\textit{MTpattern} optimizes the clustering of BIS by finding minimum number of disjoint (non-overlapping) BIS clusters such that every BIS is a member of some cluster and the exemplar BIS of any cluster has every member BIS of the same cluster in its Complete $\Omega$-Covering. This task can be broken down to a \textit{constrained optimization problem} where we need to minimize number of clusters under the constraint that every BIS should be a member of some cluster and every member of a cluster should be $\Omega$-covered by the exemplar BIS of the corresponding cluster. If there are multiple arrangements of clusters that satisfy the above constraint, then that arrangement should be chosen which minimizes net dissimilarity of all BIS with their corresponding cluster's exemplar BIS. 

\textit{MTpattern} achieves this optimization by using affinity propagation which is a relatively new clustering technique based on the concept of ``message passing" between data points (or BISs). It starts by considering all BISs as candidate exemplars and exchanges messages between every pair of BISs in every iteration till a good set of exemplars are obtained and the algorithm converges. The advantage of this technique is that it does not need number of clusters to be pre-specified and it clusters around ``exemplar" BISs \cite{frey07affinitypropagation} (members of the input set that are good representative of their corresponding cluster). This suits to the problem of behavior modeling as it is not possible to have idea about the number of underlying modes or clusters in the visiting sequences of individuals beforehand. 

 \textit{MTpattern} minimizes the number of clusters by tuning the \emph{preference} parameter of affinity propagation\footnote{``The preference of point $i$, called $p$($i$) or $s$($i$, $i$), is the \emph{a priori} suitability of point $i$ to serve as an exemplar. Preferences can be set to a common global value, or customized for every data point. High values of the preferences will cause affinity propagation to find many exemplars (clusters), while low values will lead to a small number of exemplars (clusters)" \cite{frey07affinitypropagation}}. The \emph{preference} parameter determines the granularity of the clusters. The value of \emph{preference} is usually set to the median of data points which outputs moderate number of clusters. On the other hand, it can be shown mathematically, that by setting \emph{preference} to a very large negative value (negative infinity), affinity propagation converges to a solution which outputs minimum number of clusters such that every BIS in every cluster is in the $\Omega$-Covering of the ``exemplar" BIS of the corresponding cluster. In the following equations, \textit{MTpattern} use \emph{similarity} metric instead of \emph{dissimilarity} metric where a similarity score is negative of dissimilarity score.
\begin{equation}
\begin{split}
Preference = s(S\textsubscript{i},S\textsubscript{i}) \,\forall i \,\,\epsilon\,\,(1, m)
\label{eq:pref}
\end{split}
\end{equation}
In Eq. \ref{eq:pref}, $s(S\textsubscript{i}, S\textsubscript{i})$ represents self similarity or \emph{preference} of data point (or, BIS) $i$. \textit{MTpattern} set the preference value for all BIS to the same global value which ensures that affinity propagation is not biased towards choosing any BIS as an exemplar beforehand.

\begin{equation}
\begin{split}
\begin{gathered}
Inter\_Sim = \sum_{\substack{i=1}}^{m}\sum_{\substack{c=1}}^{C}
\begin{cases}
     s(i, ex\textsubscript{c}),& s(i, ex\textsubscript{c}) \neq \infty \\
    0,              & \text{otherwise}
\end{cases}
\end{gathered}
\end{split}
\label{eq:inter_similarity}
\end{equation}

\begin{equation}
\begin{split}
\begin{gathered}
\\Preference \approx -\infty
\end{gathered}
\end{split}
\label{eq:pref_is_comp}
\end{equation}

In Eq. \ref{eq:inter_similarity}, $s(i,j)$ is the similarity score between BIS $i$ and BIS $j$ and is equal to the negative of dissimilarity score between BIS $i$ and BIS $j$. $C$ represents total number of clusters and $ex\textsubscript{c}$ represents exemplar BIS of the cluster $c$.

In Eq. \ref{eq:pref_is_comp}, \textit{MTpattern} set the value of \textit{preference} to a large negative value, much smaller than $Inter\_Sim$ which is the sum of all pair-wise similarities between similar BISs.

Affinity propagation seeks to find number of clusters that maximizes the total similarity for each cluster which is measured as the sum of the similarities between non-exemplar BISs and their exemplar BIS and the sum of preferences for selected exemplars BISs.

A formal statement of the \textit{MTpattern} optimization problem that underlies affinity propagation begins with the definition of two inputs: the similarity matrix, $s(i, j)$; and the \textit{preference}. There are two sets of decision variables associated with the optimization problem: $y$\textsubscript{c} = 1 if a BIS \textit{c} is selected as an exemplar and 0 otherwise, for 1 $\leq$ \textit{c} $\leq$ \textit{C}; and $x$\textsubscript{$ic$} = 1 if BIS \textit{i} is assigned to the cluster for which BIS \textit{j} serves as an exemplar and 0 otherwise, for 1 $\leq$ \textit{i} $\leq$ \textit{m} and 1 $\leq$ \textit{c} $\leq$ \textit{C}. The integer linear programming formulation of the problem can then be stated as follows:

\begin{equation}
Net\_Sim =  \sum_{\substack{i=1}}^{m}\sum_{\substack{c=1}}^{C} s(i, \\ ex\textsubscript{c})*x\textsubscript{$ic$}+\sum_{c=1}^{C} s(ex\textsubscript{c}, \\ ex\textsubscript{c})*y\textsubscript{c}
\label{eq:net_similarity_AP}
\end{equation}

subject to
\begin{equation}
\sum_{\substack{c=1}}^{C} x\textsubscript{$ic$}, for \, all \, 1 \leq \textit{i} \leq \textit{m} \, \textrm{and} \, ITDist(i, \\ ex\textsubscript{c}) < \Omega;
\label{eq:xic}
\end{equation}

\begin{equation}
x\textsubscript{$ic$} \leq y\textsubscript{c}, for \: all \, 1 \leq \textit{i} \leq \textit{m} \, \textrm{and} \, 1 \leq \textit{c} \leq \textit{C};
\label{eq:xic-yc}
\end{equation}

\begin{equation}
x\textsubscript{cc} = y\textsubscript{c}, for \, all \, 1 \leq \textit{c} \leq \textit{C}.
\label{eq:xcc-yc}
\end{equation}

\begin{equation}
x\textsubscript{$ic$} \in \{0,1\}, for \, all \, 1 \leq \textit{i} \leq \textit{m} \, \textrm{and} \, 1 \leq \textit{c} \leq \textit{C};
\label{eq:xic-0}
\end{equation}

\begin{equation}
y\textsubscript{c} \in \{0,1\}, for \, all \:  1 \leq \textit{c} \leq \textit{C}.
\label{eq:yc-0}
\end{equation}

The objective function (equation (5)) of the optimization problem is $Net\_Sim$, and the first and second terms on the right-hand side of the equals sign are BISs similarity and exemplar preference similarity, respectively. Constraint set (6) guarantees that each BIS is assigned to exactly one exemplar and each non-exemplar BIS in a cluster is $\Omega$-covered by the exemplar BIS of the corresponding cluster under the given constraint of local dissimilarity. Constraint set (7) ensures that a BIS is not assigned to an BIS that is not selected as an exemplar. Constraint set (8) is incorporated in the affinity propagation algorithm to ensure that, if a BIS is selected as an exemplar, then that BIS must be assigned to the cluster for which it serves as the exemplar. Finally, constraint sets (9) and (10) enforce the binary restrictions on the x\textsubscript{$ic$} and y\textsubscript{c} variables, respectively.

\begin{equation}
Net\_Sim = \sum_{\substack{i=1\\i\neq c}}^{m}\sum_{\substack{c=1\\c\neq i}}^{C} s(i, ex\textsubscript{c})+\sum_{c=1}^{C} s(ex\textsubscript{c}, ex\textsubscript{c}))
\end{equation}

\begin{equation}
\qquad\qquad =  \sum_{\substack{i=1\\i\neq c}}^{m}\sum_{\substack{c=1\\c\neq i}}^{C} s(i, \, ex\textsubscript{c})+C\times preference\\
\end{equation}

\begin{equation} 
\qquad \qquad \approx C\times preference \text{ (from Eq. \ref{eq:inter_similarity} and \ref{eq:pref_is_comp})}\qquad
\label{eq:net_similarity}
\end{equation}

\textit{MTpattern} maximizes $Net\_Sim$ similar as affinity propagation \cite{frey07affinitypropagation} does. For the $x\textsubscript{$ic$}$ = 1 and $y\textsubscript{c}$ =1, we get the Eq. \ref{eq:net_similarity} where $s$($ex\textsubscript{c}$, $ex\textsubscript{c}$) is also equal to \textit{preference} (from Eq. \ref{eq:pref}). As preference is set to -$\infty$, increasing the number of clusters ($C$) will decrease the $Net\_Sim$ value as the preference is negative and the goal is to maximize $Net\_Sim$. Therefore, a globally stable solution will minimize total number of clusters (i.e., $C$) for convergence.

Once clusters are obtained, an average of all the BISs in the cluster is taken for every unit time interval. This average gives us a probabilistic view of an individual's behavior in every time slot. This average is the manifestation of the cumulative effect of all BIS in the same cluster. Because the dissimilarity matrix is sparse, the number of messages exchanged between data points in every iteration of affinity propagation is significantly less reducing time complexity (assuming serial message passing) and space complexity of clustering.

In the Figure \ref{fig:dissimilarity_matrix}, after applying affinity propagation, we get two clusters. The exemplar BIS of first cluster is S-3 and it contains S-1, S-3, S-4, S-6 and S-7. The exemplar BIS of second cluster is S-5 and it contains S-2 and S-5. The $Net\_Sim$ (see Eq. \ref{eq:net_similarity}) is minimum in this configuration.

\section{Complexity Analysis}
In this section, we shall discuss time and space complexity for two main components of the proposed solution, i.e., dissimilarity measure and pattern discovery in detail.

\subsection{Time Complexity Analysis}

\subsubsection{Distance Measure Computation}
The time complexity of the distance measure (Algorithm 3) between two segments of \textit{N\textsubscript{s}} length, each can be given by the total number of MIN-ITDist’s calculated. Maximum number of times MIN-ITDist is executed for a pair of segment is equal to the total number of 1’s in both the sequences combined which can be 2\textit{N\textsubscript{s}} in the worst case. One instance of MIN-ITDist method runs for O($\Omega$) times. So, complexity of distance computation is 2O($\Omega$)\textit{N\textsubscript{s}}, i.e., \textit{O}(\textit{N\textsubscript{s}}$\Omega$). Since there are \textit{e}\textsuperscript{2} pairs of sequences for every segment (where \textit{e} is the total number of days), in worst case, time complexity to generate distance matrix for a time segment of length \textit{N\textsubscript{s}} is \textit{O}(\textit{e}\textsuperscript{2}\textit{N\textsubscript{s}}\textit{$\Omega$}).

The above distance matrix represents one node in the segment tree. Similar operation needs to be done for every node (every segment at every level). Time complexity for constructing distance matrix for the segment tree is equal to sum of time complexity to generate distance metric of all nodes. If the total length of root node (unsegmented sequence) is \textit{L} \textit{BIS}, height of the tree will be O(log\textsubscript{2}\textit{L}) since we divide the segment length by 2 at every level. 

Total number of leaf nodes is (2\textsuperscript{log\textsubscript{2}\textit{L}}) = \textit{L}. Time complexity for computing distance metric for all leaf nodes = \textit{O}(\textit{L}\textit{k}\textsuperscript{2}\textit{N\textsubscript{s}} \textit{$\Omega$}). Once, we compute the distance matrix for leaf nodes in the segment tree, distance matrix for non-leaf nodes in the segmentation can be computed using the distance matrix of its children nodes without any information loss, i.e., by using \textit{eBIS}. The time complexity for generating distance matrix for all non-leaf nodes at a given depth \textit{d} can be given by time complexity for computing one node at that level \textit{K}\textsuperscript{2} (distance between one pair of segment can be calculated in constant time using distance matrix of its children nodes. There are total \textit{e}\textsuperscript{2} pairs) multiplied by total number of nodes at that level O({2}\textsuperscript{\textit{d}}) = O(\textit{e}\textsuperscript{2}{2}\textsuperscript{\textit{d}}).
The total cost will be the sum of time complexity for leaf nodes and non-leaf nodes given by 

\textit{O}(\textit{L}\textit{e}\textsuperscript{2}\textit{N\textsubscript{s}}\textit{$\Omega$}) + $\sum_{\textit{d}=2^0}^{2^{log\textsubscript{2}L}} O(\textit{K}\textsuperscript{2} ×  \textit{2\textsuperscript{d}})$

$\Rightarrow$ O(\textit{Ne}\textsuperscript{2}\textit{N\textsubscript{s}}$\Omega$) + O(\textit{Lk}\textsuperscript{2})

$\Rightarrow$ O(\textit{L}\textit{e}\textsuperscript{2}\textit{N\textsubscript{s}}\textit{$\Omega$}).

Thus, creating hierarchical segments from leaf nodes does not asymptotically take more time complexity than it would take for computing distance matrix for leaf nodes alone.

\subsubsection{Pattern Discovery}
For clustering, \textit{MTpattern} uses affinity propagation. The underlying concept of this algorithm is belief propagation and the worst time complexity is 
O(\textit{r'}\textit{d\textsubscript{p}}\textsuperscript{2}) where \textit{r'} is the total number of iterations for convergence and \textit{d\textsubscript{p}} is the total number of data points. For sparse distance matrix, this time complexity is less as messages will not be passed between data points between for whom distance is not defined (or infinite). For a given distance matrix with \textit{d\textsubscript{p}} data points, time complexity is O(\textit{r'}\textit{d\textsubscript{p}}\textsuperscript{2})$\Rightarrow$ O(\textit{d\textsubscript{p}}\textsuperscript{2}).

\subsection{Space Complexity Analysis}
\subsubsection{Distance Measure Computation}
Space complexity to store distance matrix for entire segment tree is equal to the size of one node multiplied by total number of nodes. Every node contains the segment dimensions [start, end], the left and right pointer of the children nodes and the distance matrix for that segment. Segment dimensions, left and right child pointers are constant size whereas distance matrix takes O(\textit{e}\textsuperscript{2}) space where \textit{e} is number of sequences (or days). For all segments at depth d  in the segmentation tree, space complexity is O(\textit{e}\textsuperscript{2}{2}\textsuperscript{\textit{d}}). For the entire segment tree, space complexity is given by
$\sum_{\textit{len}=2^0}^{2^{log\textsubscript{2}L}} O(\textit{e}\textsuperscript{2} × {2}\textsuperscript{\textit{d}})$ $\Rightarrow$ O(L\textit{e}\textsuperscript{2}).

\subsubsection{Pattern Discovery}
The exhaustive search for all $\Omega$-coverings for a given time interval is in place so it does not require any extra space. For clustering using affinity propagation, there is need to keep in memory two messages (responsibility and availability) from every other data point, so, space complexity of single instance of affinity propagation is O(\textit{e}\textsuperscript{2}).

\section{Experiments and Analysis}
At the high-level, an effective clustering algorithm should be able to cluster similar users together and different ones separately. We evaluate our behavioral clusters quality by finding how well they capture similar users. To evaluate clustering quality, internal and external evaluation measures are used. Internal criteria are used for finding clustering quality when ground truth in the dataset is not available while, external criteria is used when ground truth is available. 

We conduct three experiments from different perspectives to evaluate and compare our proposed approach \textit{MTpattern} with baselines. We select three widely used clustering algorithms as baselines: K-means, Hierarchical Clustering (HC), and a variant of HMM, Expectation-Maximization (EM)  \cite{dempster1977maximum}. We also compare \textit{MTpattern} with one of the most popular dimensionality reduction techniques, PCA. \textbf{Experiment 1:} Evaluation and Comparison with Baselines through Internal Criteria. In Internal measure, the clustering evaluation is compared only with the result itself, i.e., to evaluate the structure of the found clusters and relationships among these clusters. Evaluation of clustering quality using internal measure is preferred in several real-world scenarios as it is not always possible to obtain ground truth with the data. While, the data labeling is expensive task. \textbf{Experiment 2:} Evaluation and Comparison with Baselines through External Criteria. In most real applications, complete knowledge of the ground truth is not available. Therefore, external measure is widely used to evaluate synthetic data. We create synthetic dataset for this purpose. \textbf{Experiment 3:} Evaluation of Distance Measure. The clustering quality is highly dependent on the distance measure used between the data objects. We compare our proposed distance measure with widely used Euclidean distance (ED) and Dynamic Time Warping (DTW). 

 In \textit{MTpattern}, the results are generated for two values of \textit{preference} for affinity propagation: \textit{median of dissimilarity scores} and \textit{$<<$median of dissimilarity scores} for different values of $\Omega$. In addition, the length of BIS (L) for any particular day is equal to 192 (1 hour = 8 * (7 mins + 30 secs), 24 hours = 192 * (7 mins + 30 secs)) as $\delta$ = 15 mins. \textit{MTpattern} uses affinity propagation based clustering on these timestamps. We perform each experiment with $preference$ parameter set to Median of data points and $-\infty$ to demonstrate efficacy of $-\infty$ preference.  The experiment is also repeated with different values of $\Omega$ ($\delta$, $2\times\delta$, $3\times\delta$ and $4\times\delta$).

\subsection{Datasets}
We use two real-world datasets and a planted (synthetic) dataset to validate the practicality of our proposed method, \textit{MTpattern} for the discovery of bahavior patterns. 

\subsubsection{WiFi dataset}
Now-a-days, the usage of Smartphones is continuously increasing all over the globe. We use the unmodified Smartphone-based user identification and tracking system (called, SmartITS) \cite{Kulshrestha} which continuously tracks MAC ids of user equipments (Smartphones/BLE tags/Bluetooth devices) in indoor-outdoor environments  seamlessly and upload the users' traces into the cloud server for the long-term analysis.

\begin{figure}[h!]
\centering
\includegraphics[scale= 0.6]{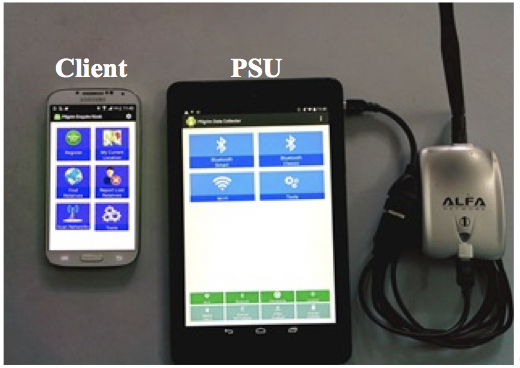}
\caption{Smartphone working as a Client (left) and Portable Sensing Unit (right).}
\label{fig:patterns2}
\end{figure}

A \textit{portable sensing unit} (PSU) collects the individuals' traces (records) in the following format $<$\textit{device id}, \textit{MAC address}, \textit{Time}, \textit{Client RSSI}, \textit{Location}$>$ where \textit{device id} is a \textit{PSU} name, \textit{MAC address} is the MAC id of the individual detected, \textit{Time} is the physical time at which an individual is detected, \textit{Client RSSI} is the signal strength of an individual to estimate the physical distance (in meters) between the \textit{PSU} and detected individual. \textit{Location} captures the GPS coordinates of the \textit{PSU}. Location comprises of latitude, longitude, and accuracy. \textit{PSUs} keep uploading the collected data at the cloud server periodically where data is stored in a time-stamped manner. Moreover, for the security and privacy reason, all MAC addresses are strictly anonymized using AES algorithm. 

Experiments are carried out in Indian Institute of Technology, Roorkee (IITR) campus. IITR is an academic and research institute in the state of Uttarakhand, India and has a 1.48 km\textsuperscript{2} campus including many objects, such as academic departments, administrative buildings, hostels, library, banks, post office, hospital, schools, canteen shops, etc. 
We set our \textit{PSUs} in the UGPC lab in the Department of Computer Science and Engineering, IITR and track all those smart devices for 3 months using \textit{SmartITS} (for which the WiFi is turned ON). In total, there are 11,853 records collected at the UGPC.

\begin{figure}[h!]
\centering
\includegraphics[scale= 0.4]{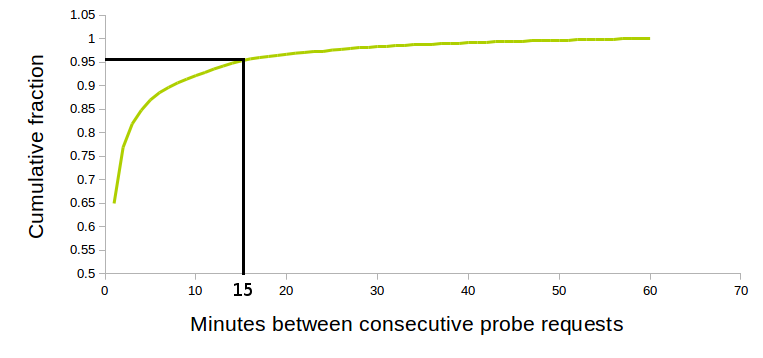}
\caption{Time  delay  between  consecutive  WiFi  probe packets}
\label{fig:probepackets}
\end{figure}


\subsubsection{Reality Mining}
Reality mining dataset was collected by MIT Media lab where 95 academic mobile phone users are tracked for approximately 9 months. We use this dataset and mine the visiting patterns of \textit{home} (using the cell tower associated with the home-ids) \cite{r2006}. Both real-world datasets have the uncertainty in the data as in WiFi dataset, sensors may fail to capture emitted packets from individuals' smart devices while reality mining dataset can loose  cell  tower  signal  or inconsequential  tower transitions  due to  dense  tower  network  and  overlapping tower range. It can be observed that both real-world datasets share the common uncertainty arising  from  inherent  temporal  variability  in any \textit{individual(s)} nature.

\subsubsection{Computation of $\delta$}
We  set  the  threshold $\delta$ equal  to  15  mins  based  on  an experimental analysis. To find the appropriate value of threshold $\delta$, we analyze duration between consecutive WiFi packets emitted by the same device in  the  direct  line  of  sight  which  are  stationary  and  within the range of the adapter. We use four android phones, one iPhone  and  one  windows  phone  for  this  experiment.  We inspect  time  delay  between  all  consecutive  WiFi  packets from  same  device  and  realize  that  for  more  than  95$\%$  of packets,  time  delay  between  the  current  packet  and  the previous packet is below 15 minutes (see Figure \ref{fig:probepackets}). Therefore, we set  the  threshold $\delta$ equal to  15  minutes with the false negative  rate  of  less  than  5$\%$. If  devices  are  not  placed  in  the  direct  line of sight  and  are  not  static  then  the  average  rate  of  false negatives  rate  for $\delta$ equal  to  15  mins  increases  to  38$\%$ due to packet loss.

In other words, threshold $\delta$ is the transmitting interval of wireless packets from client smart device in order to advertise their presence to nearby devices or actively discover the access points in its proximity. The computation of the threshold $\delta$ depends on the data collection strategy. For example, in the reality mining dataset, Bluetooth scans periodically at 5 minute intervals. While we collect all the incoming wireless packets and then preprocess them in the Time Point Sequence (PS) as shown in Figure 3. Then, we use the threshold $\delta$ to convert point sequence data into interval sequence data.

Furthermore, threshold $\delta$ determines the width of discretized bin where the width of discretized bin decides the granularity of clustering. Size of each cluster is described by granularity. Therefore, we can say that we use threshold $\delta$ to decide the size of a cluster instead of pre-specifying number of clusters.

\subsubsection{Planted (synthetic) Patterns}
Clustering efficiency can be accurately measured on synthetic datasets, since the true distribution and its modes are known. We generate different planted patterns to remove any bias that can be present in the above real-world datasets with a pre-determined temporal variability and sensor uncertainty. For simulating temporal variance, we vary the sequences' start-end in a given visiting mode/pattern so that the start/end points follow a normal distribution around a mean with 3x$\sigma$ (standard deviation) equal to 4 $\lambda$ (1 $\lambda$ = 7.5 minutes, 4 $\lambda$ = 30 minutes). When we perform clustering on planted patterns, we set $\Omega$ between 30 minutes (4 $\lambda$) and 1 hour (8$\lambda$). Note that, by definition 99.74\% of data points reside between -3$\sigma$ and 3$\sigma$, so we have significantly reduced the probability of any MIN\_ITDist being more than $\Omega$ because of temporal variation, which is inline of our assumption of a small inherent temporal variability in human actions. We set the visiting pattern with an error probability of 0.2, i.e., sensor can fail to sense the signal even though an individual was in the surrounding with a probability of 0.2 for a given \emph{IS}. The probability of erroneously failing to sense a signal over consecutive BISs reduces with a power of 0.2. So, false negative probability for 30 minutes (4$\lambda$) is 0.2\textsuperscript{4} = 0.0016 which is again very less. This dataset has only 1000 different and random patterns.

\subsection{Experiment 1 : Evaluation of Clustering Quality through Internal Criteria}
We evaluate \textit{MTpattern} using two internal measures metrics: number of clusters and accuracy score. Both metrics consider some data points as representatives of each cluster. Below, we define them formally.
\begin{enumerate}
\item Number of clusters: It is the total number of clusters achieved for the given data.  
\item Accuracy Score: It is an output pattern to correctly represent all members of a cluster. Small score means high accuracy and high score means low accuracy.
\end{enumerate}

\begin{figure*}[h!]
  \centering
  \begin{subfigure}[t]{0.3\textwidth}
    \centering
    \includegraphics[scale = 0.3]{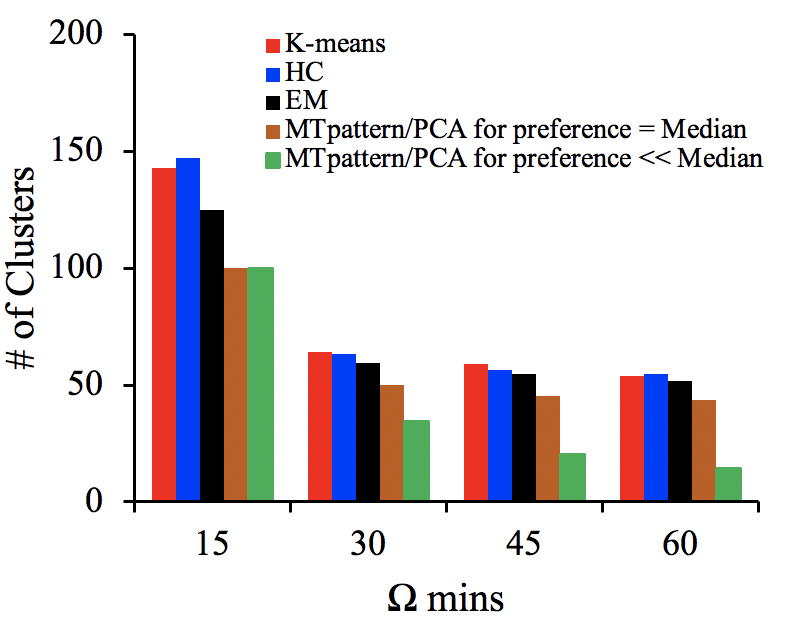}
    \caption{\label{fig:fig11}}
  \end{subfigure}
  ~
  \centering
  \begin{subfigure}[t]{0.3\textwidth}
    \centering
    \includegraphics[scale = 0.3]{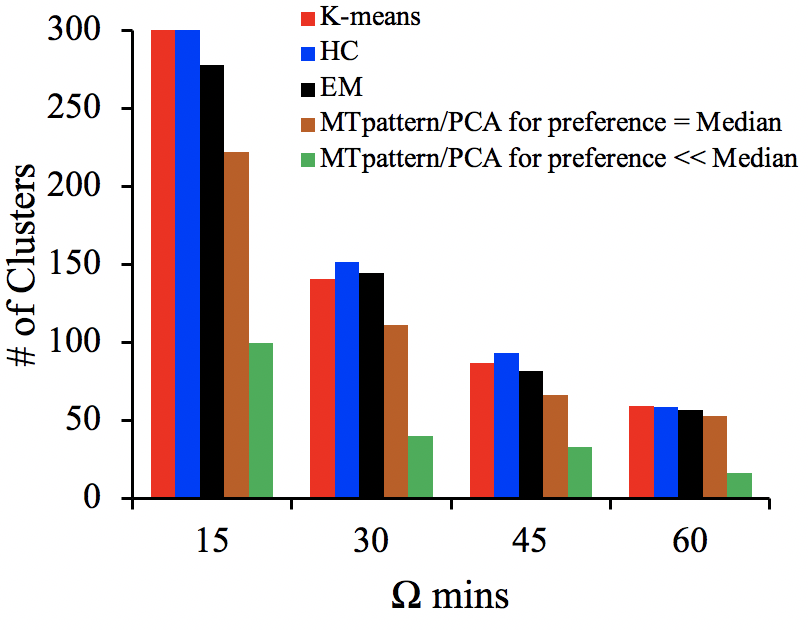}
    \caption{\label{fig:fig12}}
  \end{subfigure}
  ~
    \centering
  \begin{subfigure}[t]{0.3\textwidth}
    \centering
    \includegraphics[scale = 0.3]{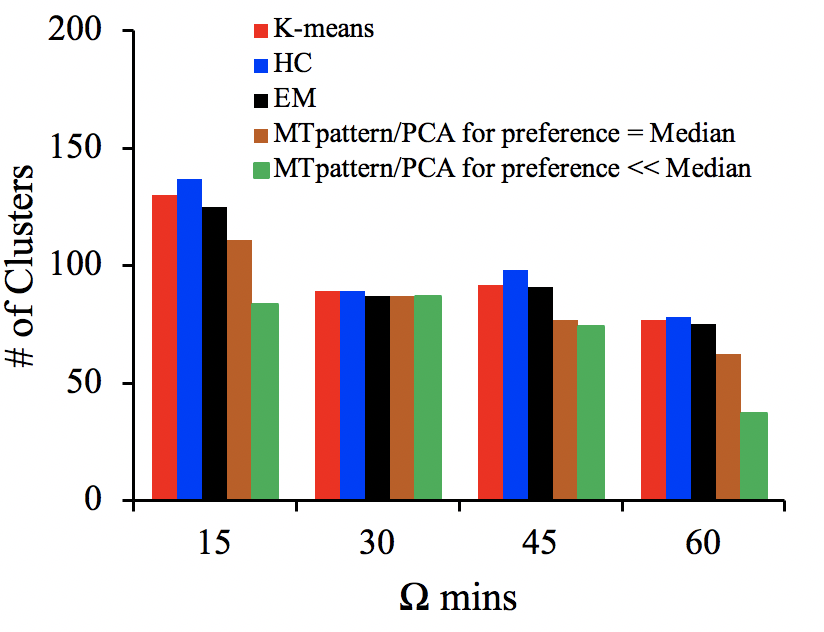}
    \caption{\label{fig:fig12}}
  \end{subfigure}
  ~
 \caption{Number of Clusters, (a) WiFi dataset (b) Reality Mining Dataset, and (c) Simulated Dataset}
 \label{fig:preProcessing1}
\end{figure*}

  We plot the total number of clusters achieved after applying the \textit{MTpattern} and $baseline$ clustering approaches. Figure \ref{fig:preProcessing1} (a), (b) and (c) show that total number of clusters achieved for all three datasets after clustering is low for our proposed technique, $MTpattern$ compared to all $baseline$ approaches. We find that for all approaches, the number of clusters decreases as the $\Omega$ increases. $MTpattern$ tunes affinity propagation to give the minimum number of clusters, while satisfying the upper bound in local dissimilarity. There is no objective way in which such minimization can be achieved for $PCA$  or for any other partition-based clustering, i.e., K-means and HC.

Since, eigenbehavior outputs eigenvectors (unit magnitude), we first convert frequent behavior patterns obtained from our proposed approach into unit vectors. We first calculate accuracy score as per the Eq. \ref{eq:ICV}. In our case, there are only two classes, \textit{present} and \textit{absent}. For every visiting sequence, we compare it with the unit vector that represents the corresponding cluster. Every correct prediction (in a unit time interval) will be rewarded and every wrong prediction will be penalized. Sum of accuracy score is taken for all segments at all segmentation levels for all individuals to represent the overall accuracy score for the techniques.

\begin{figure*}[h!]
  \centering
  \begin{subfigure}[t]{0.3\textwidth}
    \centering
    \includegraphics[scale = 0.27]{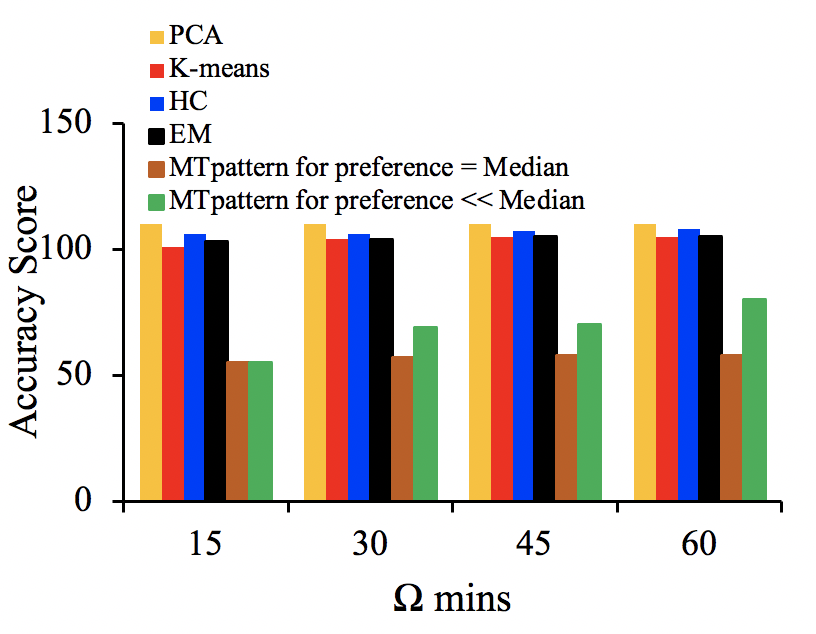}
    \caption{\label{fig:fig13}}
  \end{subfigure}
  ~
  \centering
  \begin{subfigure}[t]{0.3\textwidth}
    \centering
    \includegraphics[scale = 0.27]{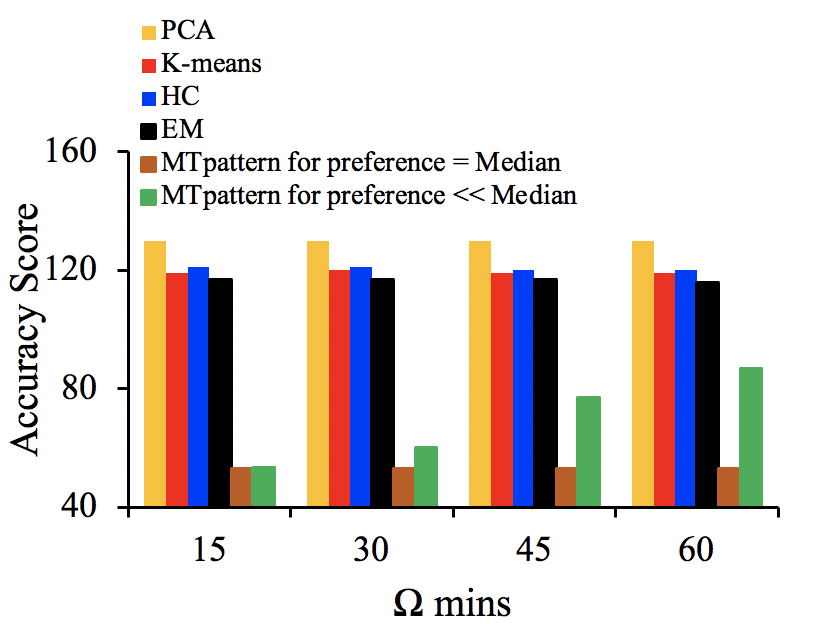}
    \caption{\label{fig:fig14}}
  \end{subfigure}
  ~
   \centering
  \begin{subfigure}[t]{0.3\textwidth}
    \centering
    \includegraphics[scale = 0.27]{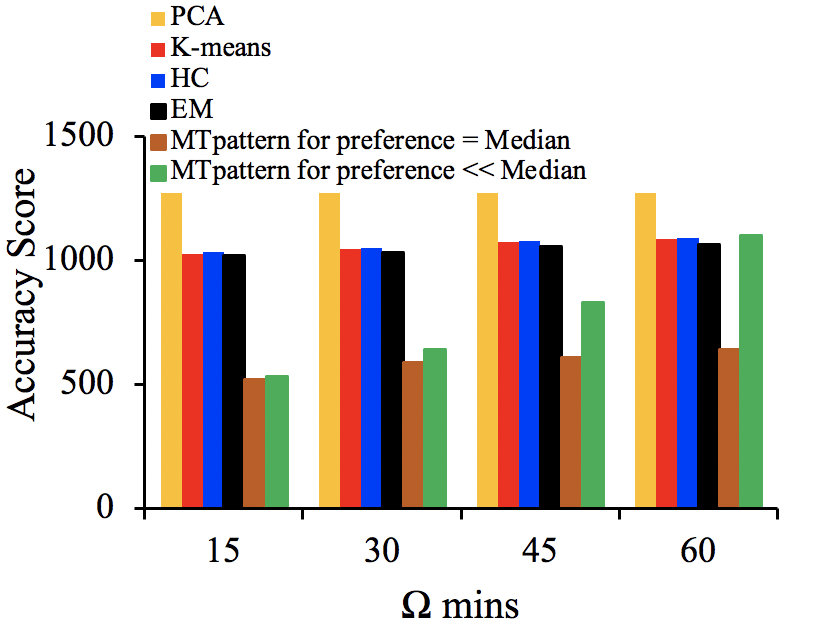}
    \caption{\label{fig:fig14}}
  \end{subfigure}
  ~ 
 \caption{Accuracy Score, (a) WiFi dataset (b) Reality Mining Dataset, and (c) Simulated Dataset}
 \label{fig:preProcessing}
\end{figure*}

\begin{equation}
\begin{split}
Accuracy\_Score = \sum_{i=1}^{e}
\sum_{j=1}^{N\textsubscript{s}}
\sum_{c=1}^{N\textsubscript{c}}
Flag\textsubscript{i}\textsubscript{j}\textsubscript{c} \times P\textsubscript{i}\textsubscript{j}\textsubscript{c}
\end{split}
\label{eq:ICV}
\end{equation}

\begin{equation}
\begin{split}
Flag\textsubscript{i}\textsubscript{j}\textsubscript{c}= 
\begin{cases}
    -1,& S\textsubscript{i}\textsubscript{j} \in c\\
    1,              & \text{otherwise}
\end{cases}
\end{split}
\end{equation}

\textit{e} is the total number of sequences or number of days. \textit{N\textsubscript{s}} is the length of a segment. \textit{N\textsubscript{c}} is total number of classes, i.e., present and absent. \textit{S}\textsubscript{i}\textsubscript{j} is the \textit{j}\textsuperscript{th} bit of sequence `\textit{i}'.
\textit{P}\textsubscript{i}\textsubscript{j}\textsubscript{c} is the value of \textit{j}\textsuperscript{th} bit of \textit{P}\textsubscript{i} (which is the frequent unit vector pattern representing the cluster which sequence `\textit{i}'  belongs to) to be of class \textit{c}. The above equation rewards correct prediction and penalizes wrong prediction.

Figure \ref{fig:preProcessing} (a), (b) and (c) show that accuracy score for all three datasets is less for our proposed technique, \textit{MTpattern} compared to $PCA$ and $baseline$ approaches. Small accuracy score shows that \textit{MTpattern} has high accuracy, i.e., \textit{MTpattern} has high cluster accuracy compared to baselines. \textit{MTpattern} takes into account local proximity between discrete time intervals unlike baseline approaches which treat every time interval as an independent dimension. 

Even though eigenbehavior representation has been used to find the human's behavior structure, it has several drawbacks: First, $PCA$ is a dimensionality reduction method which does not exploit domain-specific knowledge.  On the other hand, assigning physical meanings to the weight values in eigenvectors is challenging and can be too subjective for humans’ behavior patterns discovery as eigenvectors themselves have no well-defined physical meanings and are only used to project data to low-dimensional space. Second, the eigenbehavior representation does not consider the uncertainty in humans’ behaviors.  Third, Eigenvectors are not suitable for clustering as this process lacks theoretical support. Our proposed model, \textit{MTpattern} compensates these drawbacks.

After doing multi-modal analysis for each time slot, frequent co-occurring behavior in different time slots in a day is presented. This gives us broader view of individuals' behavior. The correlated behavior may not be serial or contiguous. E.g., a peculiar early morning behavior may lead to some different behavior at night. We owe this advantage to the segmentation of day into time slots. Early techniques, like eigenbehavior fail to capture non-contiguous co-occurring frequent behavior.

As we have already mentioned that determining  number  of  clusters  in  advance  or  maximum permissible representational error is not trivial and is often subjective. Partition-based clustering, like K-means and HC clustering algorithms and model-based clustering, like EM suffer from the above-mentioned issue. Moreover, \textit{MTpattern} puts an upper bound on local dissimilarity which is allowed for the two sequences to be similar, whereas there is no such provision in $PCA$, partition-based clustering and model-based clustering. 

\begin{figure}[!h]
\centering
\includegraphics[scale = 0.4]{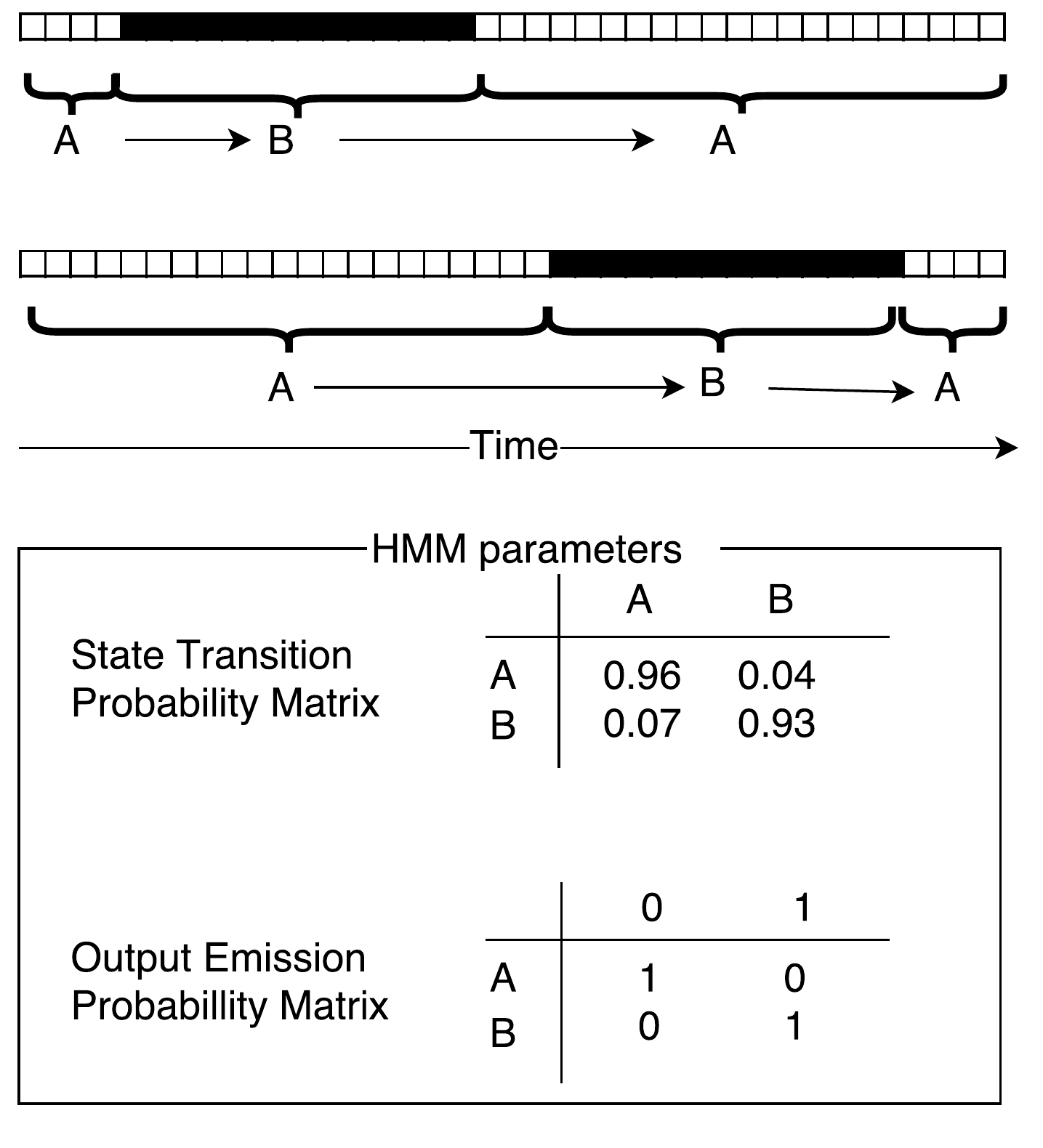}
\caption{One EM model (with 2 hidden states) fits two different binary sequences.}
\label{fig:ModelBasedClustering}
\end{figure}

On the other hand, we also observe that model-based sequence clustering methods, like HMM and EM are more sensitive to the order of events and invariant to the actual time of occurrence of the events as shown in Figure \ref{fig:ModelBasedClustering}. We fix the number of hidden states to 2\footnote{There is no golden rule to find the correct number of hidden states in HMM}. As it turns out, EM for both the cases will have same sequence of hidden states. The EM models for both the cases also has same transition and emission probability matrix. So, EM based clustering will render these two sequences indistinguishable. We also assume the observations in Figure \ref{fig:ModelBasedClustering} to be deterministic which is prerequisite for model-based clustering to work. But, real-world data often contains uncertainty and non-deterministic instances in observation as shown in Example 1 (see Figure 1).

Also, we observe through the results in Figure 10 and 11 that synthetic dataset has similar performance pattern for the proposed approach, \textit{MTpattern} and baseline clustering approaches. This gives confidence that synthetic dataset has similar data distribution as real-world datasets. Thus, synthetic dataset can be used further for evaluating clustering quality.  

Furthermore, we evaluate the time efficiency of \textit{MTpattern} for the optimal number of clusters achieved in WiFi dataset ($\Omega$ = 15 mins) w.r.to K-means, EM and Hierarchical clustering algorithms. The execution time of \textit{MTpattern}, k-means, EM and Hierarchical clustering algorithms are 9 mins, 28 mins, 16 mins, and 43 mins, respectively. \textit{MTpattern} seems to outperform methods that randomly re-sample potential centers. Finding $C$ clusters within $d\textsubscript{p}$ data points involve a search space of size $d\textsubscript{p}$ choose $C$, or [$d\textsubscript{p}$($d\textsubscript{p}$-1)($d\textsubscript{p}$-2)...($d\textsubscript{p}$-$C$+1)]/[$C$($C$-1)($C$-2)...(2)], which is polynomial in $d\textsubscript{p}$ if $C$ is within a fixed distance of 1 or $d\textsubscript{p}$, but exponential in $d\textsubscript{p}$ if $C$ grows linearly with $d\textsubscript{p}$. k-centers clustering algorithms, such as K-means depends on random initialization and performs better with smaller search space  as the initialization will be more likely to be sampled from a region near the optimal solution. Therefore, we can say that these methods work well when $C$ is close to 1 or $d\textsubscript{p}$. On the other hand, \textit{MTpattern} exploits Affinity which does not rely on random sampling. Affinity propagation exchanges messages between data points (i.e., $d\textsubscript{p}$) while considering all of them to be potential exemplars. It works better in the large-search-space situation, where $C$ is not close to 1 or $d\textsubscript{p}$.

\subsection{Experiment 2: Evaluation of Clustering Quality through External Criteria}
We evaluate \textit{MTpattern} using three external measures metrics: \textit{Purity}, \textit{Rand Index} and \textit{F-measure}. Below,  we define them formally.
\begin{enumerate}
\item Purity: It measures the ratio of the items that are in the cluster with the same class as its own.
\item Rand Index: It measures the accuracy of the clustering result in terms of percentage of decision that is correct. 
\item F-measure: It is a statistical classification measure which considers both the precision and the recall to compute the score. Precision is the number of correct results divided by the number of all returned results and the recall is the number of correct results divided by the number of results that should have been returned. 
\end{enumerate}

We calculate these measures for the simulated dataset as for this dataset, we have knowledge about the clusters and their membership. So, it is possible to compare the actual original clusters with the clusters that \textit{MTpattern} has discovered.

\begin{table}[h!]
\centering
 \begin{tabular}{||c c c c c||} 
 \hline
 Metrics & K-means & HC & EM & MTpattern \\ [0.5ex] 
 \hline\hline
Purity & 0.68 & 0.66 & 0.69 & 0.98 \\ 
 \hline
 Rand Index	& 0.64 & 0.63 &	0.65 &	0.93 \\
 \hline
 F-Measure & 0.55 &	0.53 &	0.59 &	0.92 \\ [1ex] 
 \hline
\end{tabular}
\caption{Results of External Measures on different clustering algorithms for 30 mins of $\Omega$}
\label{table:1}
\end{table}

For calculating purity, every cluster is assigned to most frequent class, then accuracy is calculated by counting the number of correctly assigned data points divided by total number of data points. High purity percentage shows that patterns are classified correctly, while low purity percentage shows wrong classification of the patterns. The purity percentage for K-means, HC and EM approach is 68\%, 66\%, 69\% for all 30 mins of $\Omega$ since these approaches are not temporally sensitive. The purity percentage of \textit{MTpattern} for preference $<<$ Median ($\approx-\infty$) is 98\% for 30 mins of $\Omega$. This result shows that \textit{MTPattern} is able to correctly cluster sequences with higher accuracy as compared to baselines. As the $\Omega$ increases, purity and number of clusters (note that number of clusters is inverse of data compression) naturally decreases and as sequences farther apart which are member of different class may be grouped in the same cluster.

We assign two items to the same cluster iff they are similar. A true positive (TP) decision assigns two similar items to the same cluster, a true negative (TN) decision assigns two dissimilar items to different clusters. But, a False Positive (FP) decision assigns two dissimilar items to the same cluster, while False Negative (FN) decision assigns two similar items to different clusters. The Rand Index (RI) measures the percentage of decisions that are correct, i.e., accuracy. But, RI gives equal weightage to FPs and FNs. Separating similar items is sometimes worse than putting pairs of dissimilar items in the same cluster. Therefore, we also use F-measure (F-Score) to evaluate the clustering quality by penalizing FNs more strongly than FPs by selecting a value $\beta > 1$.

Table 2 shows a comparison between K-means, HC, EM and MTpattern algorithms from the point of the view of purity, Rand Index (RI), and F-measure. 
Purity, RI and F-measure of \textit{MTpattern} are significantly higher than the baseline approaches which confirms a better clustering quality. 

\subsection{Experiment 3: Evaluation of Distance Measure }
There are many methods to calculate the distance information; the choice of distance measures is a critical step in clustering. It calculates the similarity between two elements and also influences the shape of the clusters. 

We compare our proposed distance measure with the well-known Euclidean distance (ED) and Dynamic Time Warping (DTW). DTW is the generalization of the ED. We define a novel distance measure between discrete time series which takes into account temporal proximity of sequences and guarantees an upper limit on the maximum variance within the cluster. We feed this sparse distance matrix into affinity propagation which finds naturally dense clusters.

Table 3 shows the F-measure for the different distance measure ED, DTW and our proposed distance measure, TDist for the Affinity Propagation clustering algorithm. Results show that our proposed distance measure, TDist outperforms ED and DTW. The reason is that the conventional distance measures used in clustering, like Euclidean distance, Manhattan distance, Jaccard distance, Kullback Leibler distance, etc., consider every time slot as an independent dimension and hence, fails to capture the temporal dynamics between neighbouring time instances.

\begin{table}[h!]
\centering
 \begin{tabular}{||c c c c||} 
 \hline
 Distance measure & 30 mins & 45 mins & 60 mins \\ [0.5ex] 
 \hline\hline
ED & 0.56 & 0.62 & 0.69 \\ 
 \hline
 DTW & 0.67 & 0.73 &	0.78 \\
 \hline
 TD & 0.92 &	0.95 &	0.97 \\ [1ex] 
 \hline
\end{tabular}
\caption{F-measure for different distance measures}
\label{table:2}
\end{table}

In addition, \emph{Dynamic Time Warping} (DTW) is widely used to find dissimilarity between temporal sequences independent of some non-linear variations between them. Unlike conventional distance measures, DTW takes into account the affinity between neighbouring time instances. But, DTW metric only takes the difference of magnitude of the two sequences after aligning them and is not suitable for binary or categorical time series with uncertainty as it does not take into account the extent of warping needed to perfectly align two sequences. Therefore, the DTW metric is not the right choice for binary or categorical time series as the DTW distance for categorical time series will always be 0 after alignment and makes it impossible to distinguish between sequences based on the extent of non-linear variation between them. Moreover, in DTW metric, it is required to map every observation to some observation of the other signal which may lead to unexpected results in case of uncertain or non-deterministic observations.



\section{Conclusion and Future Works}
In this paper, we mine temporal variable patterns from uncertain temporal data. We propose a novel approach to effectively cluster behavior patterns of individuals (named, \textit{MTpattern}) from the temporal data. We propose a dissimilarity measure, called TDist, between visiting sequences that considers only temporal distance between deterministic values. Our dissimilarity measure is sensitive to local temporal differences between sequences and restricts local temporal difference between similar sequences below a threshold. Since, it is not possible to know the number of patterns exhibited by an individual in a given segment, we employ affinity propagation, a non-parametric exemplar based clustering technique. We optimize the clustering by tuning the $preference$ value to output minimum number of clusters such that every member sequence of a cluster is similar to the exemplar sequence of the cluster. We also use segment tree data structure to pre-compute distance matrix for segments which can be in turn used to compute distance matrix for any interval. Our extensive experiments show that \textit{MTpattern} outperforms the several baseline clustering approaches. 

In future, we can correlate behavior patterns from multi-source data with various external factors to help us to understand 'humans' behavior more accurately.\\

\ifCLASSOPTIONcaptionsoff
  \newpage
\fi



\bibliographystyle{IEEEtran}
\bibliography{Bibliography}

\begin{IEEEbiography}[
{
\includegraphics[width=1in,height=1.25in,clip,keepaspectratio]{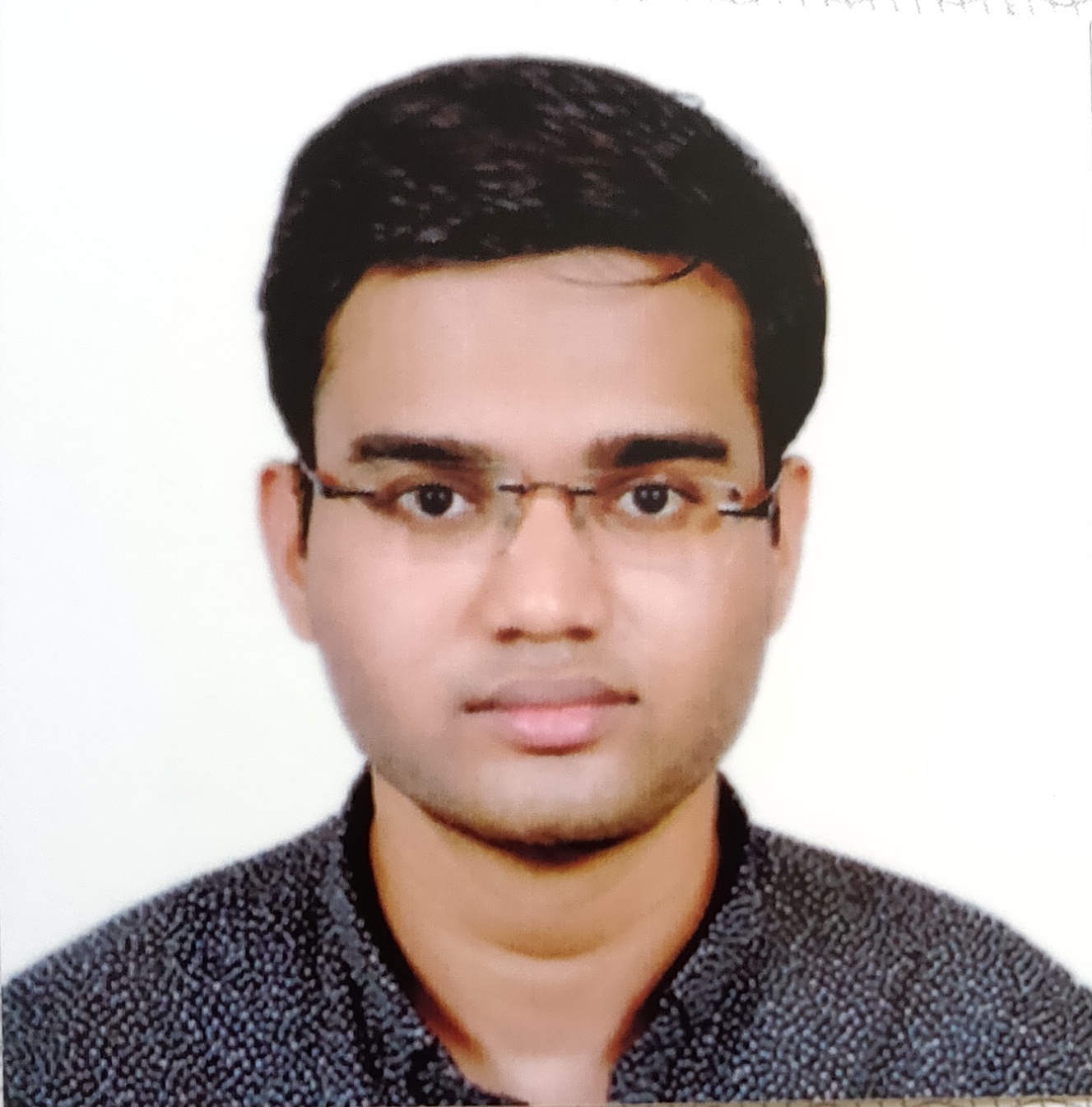}
}
]
{Rohan Kabra} received B.Tech and M.Tech degrees in CSE from Indian Institute of Technology (IIT) Roorkee in 2016, India. Currently, he is working as Senior Software Developer in Alexa Team, Amazon, USA. His research interests include Machine Learning, Data Mining and Big Data technologies.

\end{IEEEbiography}

\begin{IEEEbiography}[
{
\includegraphics[width=1in,height=1.25in,clip,keepaspectratio]{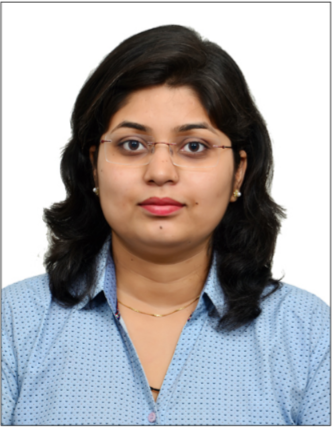}
}
]
{Divya Saxena} received the M.Tech. and PhD degrees in CSE from the IIITM, Gwalior, India and Indian Institute of Technology (IIT) Roorkee, India in 2012 and 2017, respectively. Currently, she is working as Research Assistant Professor in the Department of Computing, The Hong Kong Polytechnic University, Hong Kong. She has also worked as Postdoc and Research Fellow, in the Department of Computing and University Research Facility in Big Data Analytics (UBDA), The Hong Kong Polytechnic University, Hong Kong. Her research interests include Generative Adversarial Networks (GANs), Spatio-temporal data mining, Image-to-Image Translation and Big data analytics. She is a member of ACM and IEEE.

\end{IEEEbiography}

\begin{IEEEbiography}[
{
\includegraphics[width=1in,height=1.25in,clip,keepaspectratio]{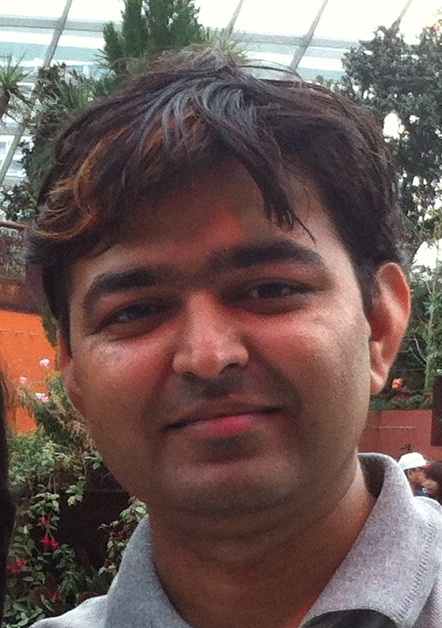}
}
]
{Dhaval Patel} received the M.Tech. and PhD degrees in Computer Science from the IIT, Kharagpur, India and NUS Singapore, in 2006 and 2011, respectively.  Currently, he is working as a Research Staff Member, IBM Thomas J. Watson Research Center, Yorktown Heights, NY USA. His research interests include Data Mining, Text Mining, Natural Language Processing. He is a Senior member of IEEE.
\end{IEEEbiography}

\begin{IEEEbiography}[
{
\includegraphics[width=1in,height=1.25in,clip,keepaspectratio]{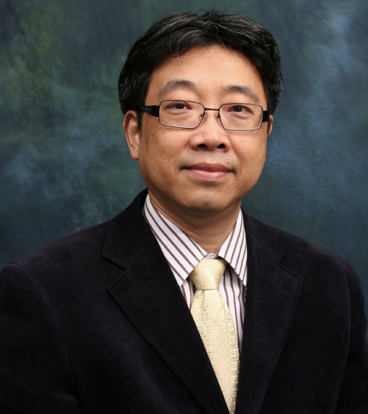}
}
]
{Jiannong Cao} received the MSc and PhD degrees in Computer Science from Washington State University, Pullman, Washington, in 1986 and 1990, respectively. He is currently the Chair Professor in the Department of Computing and Associate Director of the University Research Facility in Big Data Analytics (UBDA), The Hong Kong Polytechnic University, Hong Kong. His research interests include parallel and distributed computing, wireless sensing and networks, pervasive and mobile computing, and big data and cloud computing. He is a fellow of the IEEE and ACM Distinguished member. 
\end{IEEEbiography}

\end{document}